\documentclass[11pt]{article}
\usepackage[numbers]{natbib}
\usepackage{packages}
\usepackage{statistics}
\usepackage{statistics-macros}
\usepackage{manyfoot}

\definecolor{innerboxcolor}{rgb}{.9,.95,1}
\definecolor{outerlinecolor}{rgb}{.6,0,.2}
\definecolor{outerlinecoloreb}{rgb}{0,1,0}
\definecolor{innerboxcoloreb}{rgb}{1,1,1}

\newcommand{\dottedline}[1]{%
  \multicolumn{1}{c}{\dotfill} & \multicolumn{1}{c}{\dotfill}\\
}

\DeclareNewFootnote{A}
\DeclareNewFootnote{B}
\DeclareNewFootnote{C}

\let\footnoteR\footnoteB
\let\footnote\footnoteA

\begin{document}


\begin{center}
  {\huge Towards Out-Of-Distribution Generalization: A Survey} \\
  \vspace{.5cm}
  {\Large
    Jiashuo Liu\footnoteR{Equal contribution}, Zheyan Shen$^{*}$, Yue He, Xingxuan Zhang, Renzhe Xu, Han Yu, Peng Cui\footnoteC{Corresponding Author}} \\
  \vspace{.2cm}
  {\large Department of Computer Science and Technology}  \\
  \vspace{.2cm}
  {\large Tsinghua University}  \\
    \vspace{.2cm}
  \texttt{liujiashuo77@gmail.com, shenzy13@qq.com, cuip@tsinghua.edu.cn}
\end{center}

\begin{abstract}
Traditional machine learning paradigms are based on the assumption that both training and test data follow the same statistical pattern, which is mathematically referred to as Independent and Identically Distributed ($i.i.d.$). 
However, in real-world applications, this $i.i.d.$ assumption often fails to hold due to unforeseen distributional shifts, leading to considerable degradation in model performance upon deployment. 
This observed discrepancy indicates the significance of investigating the Out-of-Distribution (OOD) generalization problem.
OOD generalization is an emerging topic of machine learning research that focuses on complex scenarios wherein the distributions of the test data differ from those of the training data. 
This paper represents the first comprehensive, systematic review of OOD generalization, encompassing a spectrum of aspects from problem definition, methodological development, and evaluation procedures, to the implications and future directions of the field. 
Our discussion begins with a precise, formal characterization of the OOD generalization problem. 
Following that, we categorize existing methodologies into three segments: unsupervised representation learning, supervised model learning, and optimization, according to their positions within the overarching learning process. 
We provide an in-depth discussion on representative methodologies for each category, further elucidating the theoretical links between them. 
Subsequently, we outline the prevailing benchmark datasets employed in OOD generalization studies. 
To conclude, we overview the existing body of work in this domain and suggest potential avenues for future research on OOD generalization. 
A summary of the OOD generalization methodologies surveyed in this paper can be accessed at \url{http://out-of-distribution-generalization.com}.
\end{abstract}

\section{Introduction}\label{sec:introduction}
Contemporary machine learning methodologies have demonstrated their superior proficiency across various domains such as natural language processing, computer vision, recommendation systems, etc. 
While these techniques have been observed to exceed human-level performance under controlled experimental conditions, a growing body of research has underscored the susceptibility of machine learning models to data distribution shifts. 
The costs of such errors vary substantially across different applications. 
While minor inconveniences, such as a suboptimal movie recommendation or a misclassified image, are generally tolerable, slight inaccuracies in high-stakes domains such as healthcare or autonomous driving can cause catastrophic consequences. 
Consequently, the exploration of Out-of-Distribution (OOD) generalization has emerged as a pressing concern in both academic and industrial fields, with a view to enhancing the robustness and reliability of intelligent systems across diverse real-world scenarios.

Despite the importance of OOD generalization, conventional supervised learning techniques cannot be straightforwardly applied to resolve it. 
From a theoretical standpoint, the fundamental assumption underpinning classic supervised learning is that of Independent and Identically Distributed ($i.i.d.$) data, postulating that the training and test datasets originate from the same distribution. 
However, this assumption is systematically violated in OOD generalization scenarios due to inevitable distributional shifts, rendering classical learning theory inadequate.
From an empirical perspective, conventional supervised learning approaches typically focus on minimizing average training errors, greedily incorporating all correlations within the data to improve predictive accuracy. 
Although this strategy has proven effective in $i.i.d.$ settings, it is detrimental to model performance under distributional shifts, as not all correlations persist in unfamiliar test distributions.
Numerous studies \citep{duchi2018learning, arjovsky2019invariant, creager2020environment, shen2020stable, hrm} demonstrate that, when confronted with severe distributional shifts, models optimized purely based on average training errors perform poorly, often proving inferior even to random guesses. 
These observations underline the urgent need for tailored methodologies to address OOD generalization problems effectively.

Addressing the Out-of-Distribution (OOD) generalization problem necessitates the resolution of several pivotal issues. 
First, a formal characterization of the distributional shifts is required, given that training and test data can originate from different distributions. 
This issue remains largely unresolved in the OOD generalization literature, with various methodological branches adopting distinct approaches to model the potential test distribution. 
Causal learning techniques \citep{peters2016causal, buhlmann2018invariance} formulate training and test distributions using causal structures, with distributional shifts largely attributed to interventions or confounding factors.
Invariant learning methodologies \citep{muandet2013domain, ganin2015unsupervised, ganin2016domain, li2018domain, arjovsky2019invariant}, on the other hand, primarily focus on real-world scenarios, leveraging data collected across diverse domains. 
Stable learning methods \citep{kuang2018stable, kuang2020stable, shen2020stable} introduce distributional shifts through selection bias. 
Second, designing an algorithm with robust OOD generalization performance is a prevalent research focus. 
This endeavor has given rise to multiple branches of methodologies, each with distinct research objectives, including unsupervised representation learning, supervised model learning, and optimization techniques.
Third, evaluating the OOD generalization performance of various methods also poses a significant challenge. 
This is due to the need for specific datasets as well as evaluation metrics, as traditional benchmarks for the $i.i.d.$ setting are not applicable. 
This situation further demonstrates the need for curated datasets and evaluation frameworks.

In this paper, we aim to deliver a systematic and comprehensive survey of research undertakings in the realm of Out-of-Distribution (OOD) generalization. 
Our survey adopts an expansive view of the OOD generalization problem, encompassing all stages from its formal definition and methodological approaches, to its evaluation, implications, and prospective directions. 
To our knowledge, this paper represents the first effort to examine OOD generalization in such an extensive, holistic manner.
While previous research efforts have addressed related topics—\citet{wang2021generalizing, zhou2021domain} primarily focus on domain generalization, and \citet{Ye2021b} discuss evaluation benchmarks for OOD generalization—these works each contribute a piece of the broader OOD generalization puzzle. 
In contrast, our work cohesively integrates these disparate elements in a clear and succinct manner. 
Specifically, we classify existing methods into three categories, based on their position in the overall learning pipeline, and elaborate on the theoretical connections between different methods through the perspective of causality. 
To further facilitate future research in OOD generalization, we also provide a comprehensive survey of datasets to evaluate learning methods under distribution shifts.

The structure of this paper is organized as follows. In Section \ref{sec:problem}, we formulate the OOD generalization problem, elucidate its relationship with existing research areas, and propose a categorization of methods. Sections \ref{sec:representation}, \ref{sec:causal}, and \ref{sec:optimization} respectively detail the representative methods of each category. Section \ref{sec:theory} offers theoretical connections and insights between different methods, while Sections \ref{sec:benchmark} and \ref{sec:fairness} summarize applicable benchmarks for OOD generalization and its potential implications. Finally, we conclude this paper in Section \ref{sec:conclusion}, suggesting promising directions for future research.

\section{Problem Definition and Categorization of Methods}
\label{sec:problem}

In this section, we first formalize the overarching Out-of-Distribution (OOD) generalization problem and illustrate its similarities and differences with the classic Independent and Identically Distributed ($i.i.d.$) learning problem. 
We then proceed to explore several research domains related to OOD generalization, including Domain Adaptation, Domain Generalization, Federated Learning, and Out-of-Distribution Detection. 
Finally, we classify existing methodologies that address OOD generalization into distinct categories based on their respective positions within the entire learning pipeline.

\subsection{Problem Definition}
Let $\mathcal{X}$ be the feature space and $\mathcal{Y}$ the label space.
A parametric model is defined as $f_\theta: \mathcal{X} \to \mathcal{Y}$, which serves as a mapping function from original features to the label with learnable parameter $\theta$.
A loss function $\ell: \mathcal{Y} \times \mathcal{Y} \to \mathbb{R}$, which measures the distance between predictions and ground-truth labels.
The classic supervised learning problem is defined as \Cref{pro:supervised_learning}.

\begin{definition}[Supervised Learning]
\label{pro:supervised_learning}
Given a set of $n$ training samples of the form $\{(x_1,y_1),\dots,(x_n,y_n)\}$ which are drawn from training distribution $P_{tr}(X,Y)$, a supervised learning problem is to find an optimal model $f_\theta^*$ which can generalize best on data drawn from test distribution $P_{te}(X,Y)$:
\begin{equation}
\label{equ:supervised_learning}
	f_\theta^* := \arg\min_{f_\theta} \mathbb{E}_{X,Y\sim P_{te}}[\ell(f_\theta(X),Y)].
\end{equation}
\end{definition}

Traditional learning algorithms typically assume that both the training and test samples are Independent and Identically Distributed ($i.i.d.$) realizations from a shared underlying distribution, namely, $P_{tr}(X, Y) = P_{te}(X,Y)$. 
Based on this assumption, the Empirical Risk Minimization (ERM) framework \cite{vapnik1991principles}, which seeks to minimize the average loss on training samples, is capable of yielding an optimal model that successfully generalizes to test distributions \cite{vapnik1999an}. 
Specifically, ERM seeks to minimize the following objective:
\begin{equation}
	\label{eqn:erm}
	\mathcal{L}_{\text{ERM}}(\theta) := \frac{1}{n} \sum \limits_{i=1}^n \ell(f_\theta(x_i),y_i).
\end{equation}
The admirable properties provided by the $i.i.d.$ assumption have served as a strong foundation for the development of numerous learning models over the past few decades. 

\paragraph{Out-of-Distribution Generalization Problem}
In real-world scenarios, the test distribution upon which a model is deployed may diverge from the training distribution \cite{quinonero2009dataset}, that is, $P_{tr}(X,Y)\neq P_{te}(X,Y)$. 
This distribution shift could be attributed to various factors, such as the temporal or spatial evolution of data, or the sample selection bias inherent in the data collection process, which render the problem more complex than the $i.i.d.$ learning scenario. 
Moreover, the test distribution that one may encounter is typically unknown due to the nature of applications like stream-based online scenarios, wherein test data are generated in the future. 
In summary, the general Out-of-Distribution (OOD) generalization problem can be defined as a specific instance of the supervised learning problem wherein the test distribution $P_{te}(X,Y)$ diverges from the training distribution $P_{tr}(X,Y)$ and remains unknown during the training phase. 

There are multiple taxonomies for the OOD generalization problem \citep{lipton2018detecting, ScholkopfJPSZM12, plex, tiffany}. 
This survey primarily focuses on a popular categorization, which classifies distribution shifts as either covariate shifts (changes in the marginal distribution $P_X$) or concept shifts (changes in the conditional distribution $P_{Y|X}$), and provides a systematic review of various methodological approaches addressing the OOD generalization problem. 
Reviews of some specific distribution shifts have been developed relatively independently and can be found in well-established surveys \citep{NEURIPS2020_219e0524, gama2014survey}.

\subsection{Categorization of Methods}

To address the challenges posed by unknown distribution shifts, a significant number of efforts have been dedicated to out-of-distribution generalization, resulting in a vast array of relevant methods. 
The adopted techniques vary extensively, ranging from causality to robustness, and from structure-based to optimization-based strategies.
However, to the best of our knowledge, little effort has been made to systematically and comprehensively survey these diverse methodologies within the broader context of OOD generalization, as well as elucidating the differences and interconnections between these works. 
In this paper, we aim to bridge this gap by reviewing the related methods of OOD generalization.

Broadly speaking, the supervised learning problem, as defined in \Cref{equ:supervised_learning}, can be divided into three relatively independent components: 
(1) The representation of features $X$ (e.g., $g(X)$);
(2) The mapping function $f_\theta(X)$ from features $X$ (or $g(X)$) to the label $Y$, which is generally also known as the model or inductive bias; 
(3) The optimization objective.
Based on this learning pipeline, we classify existing methods into three categories, according to their respective positions in the pipeline:
\begin{itemize}
    \item \textbf{Unsupervised Representation Learning for OOD Generalization} includes unsupervised domain generalization and disentangled representation learning, which exploit the unsupervised representation learning techniques to initialize a better representation for downstream OOD generalization tasks.
    \item \textbf{Supervised Model Learning for OOD Generalization} includes invariant representation learning, training strategy, causal learning, invariant risk minimization, stable Learning, and heterogeneity-aware invariant learning, which design various model architectures and learning strategies to achieve OOD generalization.
    \item \textbf{Optimization for OOD Generalization} includes distributionally robust optimization and other variants, which directly formulate the objective of OOD generalization and mainly focus on robust optimization with theoretical guarantees for OOD optimality.
\end{itemize}
Within each primary category, we have established numerous sub-categories based on differing technical approaches and any additional information prerequisites.

\subsection{Discussion on Related Topics}
We then discuss several research topics related to the OOD generalization problem. 

\paragraph{Domain Adaptation \& Generalization}
A field related to OOD generalization is domain adaptation, which assumes the accessibility of the testing distribution, whether labeled $P_{te}(X,Y)$ or unlabeled $P_{te}(X)$. 
Domain adaptation can be viewed as a particular instance of OOD generalization where there is some prior knowledge of the test distribution.
Under such conditions, domain adaptation can avail itself of theoretical guarantees \cite{ben2010theory} which maintain the optimality of the trained model in test scenarios. 
The detailed exploration of domain adaptation methods is beyond the scope of this paper; interested readers may refer to well-established studies in this area~\cite{patel2015visual, pan2009survey, Sun2015, zhuang2021a, csurka2017domain, long2013transfer}.

Over the past decade, domain generalization (DG), a popular branch of methodology, has rapidly garnered research attention \cite{blanchard2011generalizing}. 
By assuming the heterogeneity of the training data, DG methods utilize additional domain (also known as environment) labels to learn an invariant model that can generalize to unseen and shifted test data. 
Supported by a number of high-quality benchmarks, domain generalization studies are primarily conducted in the field of computer vision tasks. 
In this paper, in order to provide balanced content from various fields, we focus on the more general out-of-distribution generalization problem and only introduce a select number of typical DG methods. 
For a comprehensive introduction to the domain generalization problem itself, one may refer to specific DG surveys \cite{zhou2021domain, wang2021generalizing}.

\paragraph{Federated learning}
Federated learning (FL), raised by \citet{mcmahan2017communication}, addresses scenarios where multiple entities (clients) collaborate to solve a machine learning problem under the coordination of a central server or service provider~\citep{kairouz2019advances}. 
Over recent years, there has been widespread interest in various aspects of FL, including communication-efficient learning, model ensemble, compression integration, system heterogeneity, data heterogeneity, personalization, and privacy. 
We direct readers to \citep{kairouz2019advances,wang2021field} for a more comprehensive survey.

Among these aspects, data heterogeneity is most closely related to OOD generalization problems. 
Both OOD and FL problems assume data heterogeneity within their training datasets. 
In OOD generalization problems, data heterogeneity is leveraged by models to infer invariant models. 
In FL, data is distributed across clients and is statistically heterogeneous as the training samples on clients may come from different distributions. 
Various assumptions \citep{gorbunov2021local,karimireddy2019scaffold,koloskova2020unified,li2019convergence} are made in regard to data heterogeneity to guarantee the performance of FL models, which is assessed as the expected utility across all clients.
The key difference between FL and OOD lies in the mode of evaluation. 
In FL, while distributions on various clients differ, researchers assume that a distribution exists among the clients, and the clients in the training and test datasets are independently drawn from this distribution. 
Consequently, the training and test datasets are \textit{i.i.d.} in a certain sense in FL. 
In contrast, the testing distribution in OOD generalization problems remains unknown.

\paragraph{Out-of-Distribution Detection}
Out-of-Distribution (OOD) detection~\citep{DBLP:journals/corr/abs-2110-11334}, aims to identify and reject unfamiliar objects not encountered during training to ensure reliability (e.g., to forward them to experts for safe handling). 
Unlike OOD generalization tasks, which primarily focus on performance under distribution shifts, the OOD detection community~\citep{DBLP:conf/nips/RenLFSPDDL19,DBLP:conf/iclr/LiangLS18,DBLP:conf/iclr/HendrycksMD19,DBLP:conf/nips/ThulasidasanCBB19} concentrates more on detecting samples from unseen distributions. 
Some works in this area also pay attention to test samples with non-overlapping labels (i.e., new classes) compared to training data.

Another related topic is open set classification, which aims to directly recognize unknown categories in test data~\citep{jia2020deep, geng2020recent}. This also differentiates from OOD generalization, where the label space is shared between the training and test data.

\vspace{0.2in}

In the following sections, we provide a comprehensive and detailed review of OOD generalization methods corresponding to the above order and discuss their differences and theoretical connections.

\section{Unsupervised Representation Learning}
\label{sec:representation}

In this section, we review methods that concentrate on unsupervised representation learning, which primarily include unsupervised domain generalization and disentangled representation learning. 
These methods either independently learn domain-agnostic features, or they employ pre-existing human knowledge to structure and regulate the representation learning process.
By doing so, they ensure that the learned representation possesses certain attributes that may facilitate out-of-distribution generalization.

\subsection{Unsupervised Domain Generalization}
Learning a discriminative and robust representation across diverse distributions, particularly with limited labeled data, can serve as the foundation for out-of-distribution (OOD) generalization ability~\citep{pmlr-v139-mahajan21b, Zhang_2022_CVPR}.
Prior to this, the issue of initializing pre-trained weights for OOD generalization has remained a crucial but often overlooked aspect. 
In the realm of computer vision, it is conventional to initialize with ImageNet pre-trained weights for OOD generalization. 
However, this can introduce significant bias. 
For instance, the "real" domain in DomainNet \cite{zhao2019multi} and the "photo" domain in PACS \cite{li2017deeper} share a similar distribution with ImageNet, while other domains exhibit distinct shifts. 
Consequently, such initialization can be viewed as pretraining on one of the source domains.
Furthermore, for datasets like NICO${++}$ \cite{he2021towards, zhang2022nico++}, where image contexts are deemed as domains, ImageNet provides additional knowledge of numerous contexts which could lead to the leakage of test domains~\citep{rethinking}.

Addressing these challenges, \citet{pmlr-v139-mahajan21b} and \citet{Zhang_2022_CVPR} introduce the concept of unsupervised domain generalization (UDG). 
This approach aims to learn generalizable models using unlabeled data, while simultaneously analyzing the effects of pre-training on OOD generalization. 
Recently, numerous self-supervised learning methods have shown promising results, using large-scale unlabeled data to learn potent representation spaces \cite{chen2020improved,he2020momentum,chen2020simple,grill2020bootstrap}. 
However, these methods cannot directly tackle the OOD generalization problem, as the learned representation space contains domain-specific features used to discriminate negative samples. 
These features can be unhelpful or even detrimental to downstream tasks \cite{zhang2021deep}. 

To address this, \citet{Zhang_2022_CVPR} propose DARLING, a method that outperforms the ImageNet pre-trained approach using significantly less unlabeled data. This indicates a promising direction for model initialization for OOD generalization. In follow-up work, \citet{harary2021unsupervised} suggest learning an auxiliary bridge domain along with a set of mappings from training domains to semantically align all domains. Another work \citep{yang2022domain} utilizes Masked Auto-Encoders \citep{he2021masked} to further enhance unsupervised representation learning. Other studies \citep{liao2020deep,zhou2021semi} also discuss semi-supervised learning approaches for the OOD generalization problem.

\subsection{Disentangled Representation Learning}
Disentangled representation learning aims to learn representations where distinct and informative factors of data variation are separated~\cite{bengio2013representation,locatello2019challenging}. 
This is considered a characteristic of high-quality representation and can potentially benefit out-of-distribution generalization.
The most prevalent methods for achieving disentanglement are based on Variational Autoencoders (VAE~\cite{higgins2016beta,kim2018disentangling}). 
These are implemented in an entirely unsupervised manner within a single environment, without additional information. 
These methods prioritize both interpretability and sparsity. 
Here, sparsity refers to the idea that small changes in distribution typically manifest in a sparse or localized manner within the disentangled factorization~\cite{scholkopf2021toward}.

\paragraph{No Additional Information} $\beta$-VAE~\cite{higgins2016beta} introduces an extra hyperparameter $\beta$ into vanilla VAE objective function, making a trade-off between latent bottleneck capacity and independence constraints, thus encouraging the model to learn more efficient representations. 
The objective function of $\beta$-VAE is as follows: 
\begin{equation}
	\mathcal{L}=\mathbb{E}_{q(z|x)}[\log p(x|z)] - \beta \text{KL}(q(z|x)\|p(z))
\end{equation}
where $z$ represents the latent representation, $x$ denotes the observed data, $p(z)$ symbolizes the prior distribution of latent factors, $p(x|z)$ indicates the decoding distribution, and $q(z|x)$ is the encoding posterior distribution. 
When $\beta$ is set to $1.0$, this formulation reduces to vanilla VAE. 
By appropriately tuning $\beta$, the $\beta$-VAE can learn disentangled representations from data in an unsupervised manner.

FactorVAE~\cite{kim2018disentangling} adds the term of Total Correlation into the objective function, which is formulated as the KL-divergence between marginal posterior $q(z)$ and its corresponding factorized distribution $\Bar{q}(z)$: 
\begin{equation}
	\mathcal{L}=\mathbb{E}_{q(z|x)}[\log p(x|z)] - \text{KL}(q(z|x)\|p(z)) - \gamma \text{KL}(q(z)\|\Bar{q}(z))
\end{equation}
where $\Bar{q}(z):=\prod_{j=1}^{d}q(z_j)$. 
This formulation encourages independence for the posterior latent representation. 
Since the Total Correlation term cannot be computed directly, an extra discriminator is added for density ratio estimation.

More recently, despite the popularity of VAE-based methods without contextual information, \citet{locatello2019challenging} challenge some common assumptions of unsupervised disentangled representation learning (e.g., independence of latent factors). 
This brings back the need for additional information to the attention of researchers. 
It also questions whether disentanglement can improve downstream task performances, inspiring later works to take downstream tasks into consideration, OOD generalization performance included.  

Among these works, a new category of disentangled representation learning arises, i.e. causal representation learning. 
Similar to conventional disentangled representation learning, causal representation learning aims to learn variables in the causal graph with the aid of auxiliary annotations.
Further, causal representation can be viewed as the ultimate goal of disentanglement, which satisfies the informal definition of disentangled representation in terms of interpretability and sparsity.
With the learned causal representation, one can capture the latent data generation process, which can help to resist the distributional shifts induced by interventions.

CausalVAE~\cite{yang2021causalvae} combines the linear Structural Causal Model (SCM) into the VAE model to endow the learned latent representation with causal structure. 
Specifically, the causal structure is depicted by an adjacency matrix $A$ as:
\begin{equation}
	z=A^T z + \epsilon = (I-A^T)^{-1}\epsilon,\quad \epsilon \sim \mathcal{N}(0, I)	
\end{equation}
where $\epsilon$ represents the exogenous factors. 
In practice, a mild nonlinear function $g_i$ is introduced for stability as $z_i=g_i(A_i\cdot z; \eta_i) + \epsilon_i$.
Further, extra labels $u$ of latent causal variables are leveraged in CausalVAE, which gives the objective function as:
\begin{equation}
	\mathcal{L}=-\text{ELBO}+\alpha \text{DAG}(A) + \beta l_u + \gamma l_m
\end{equation}
where ELBO represents the Evidence Lower Bound, $\text{DAG}(A)$ the Directed Acyclic Graph (DAG) constraint, $l_u=\mathbb{E}_{q_{\mathcal{X}}}||u-\sigma(A^T u)||_2^2$ measures how well $A$ describes causal relations among labels, and $l_m=\mathbb{E}_{z \sim q_{\phi}} \Sigma_{i=1}^n||z_i-g_i(A_i \cdot z; \eta_i)||_2^2$ measures how well $A$ describes causal relations among latent codes.
$q_{\mathcal{X}}$ is the empirical data distribution and $q_{\phi}$ the approximate posterior distribution.

Moving a step on, DEAR~\cite{shen2020disentangled} incorporates nonlinear SCM with a bidirectional generative model and assumes the known causal graph structure and extra supervised information of latent factors.
The objective function is given as:
\begin{equation}
	\mathcal{L}(E, G, F)=\mathcal{L}_{gen}(E, G, F) + \mathcal{L}_{sup}(E)    
\end{equation}
where $E,G$ denotes the encoder and generator, respectively. 
The first part $\mathcal{L}_{gen}(E, G, F)=\text{KL}(q_E(x,z), p_{G,F}(x,z))$ resembles the VAE loss. 
The difference lies in the prior distribution of $z$. 
In DEAR, this prior is generated by the nonlinear SCM, while in vanilla VAE, it is simply a factorized Gaussian. 
The second part is $\mathcal{L}_{sup}(E)=\mathbb{E}_{x,y}\text{CE}(\Bar{E}(x),u)$, where CE is the cross entropy loss function, $\Bar{E}$ represents the deterministic part of $E$ and $u$ the extra labels.

\paragraph{Require Additional Information} Apart from these fully unsupervised disentangled representation learning methods, there are works utilizing additional information toward disentanglement. 
The additional information, in the broad sense, includes not only environment labels or domain labels but also auxiliary annotations such as additional information related to latent causal variables. 
\citet{reed2014learning} propose disentangling Boltzmann machines by incorporating partial labels to construct corresponding data pairs. 
\citet{zhu2014multi} and \citet{yang2015weakly} both explicitly take ground truth transformed images and transformations as auxiliary signals. The former is based on a directed graphical model optimized in a style of EM algorithm, while the latter employs RNN to capture longer-term dependency for multiple transformation steps, inspired by mental experiments on human beings. 
Meanwhile, \citet{kulkarni2015deep} implicitly take advantage of contextual information by enforcing specific latent variables held fixed, leaving only other properties varying in a single training batch. 
Recently, \citet{DBLP:conf/cvpr/ZhangZLWSX22} propose a primal-dual algorithm for joint representation disentanglement and domain generalization, which shows the potential of disentanglement to enhance generalization ability.

In addition to the above disentangled representation learning methods, with or without additional information, there exist discussions and explorations on how disentangled representation can benefit OOD generalization. 
\citet{2020arXiv200607796L} take advantage of causal ordering information and conduct some quantitative extrapolation experiments, finding that the learned disentangled representation fails to extrapolate to unseen data, 
while \citet{trauble2021disentangled} and \citet{dittadi2020transfer} empirically verify the ability to generalize under OOD circumstances. 
\citet{reddy2021causally} propose to utilize bounding box information to achieve disentanglement and come up with a new dataset CANDLE that can be applied in multiple settings including OOD generalization tasks. 
\citet{lachapelle2023synergies} prove that disentangled representations incorporated with sparse task-specific predictors could improve generalization.
Overall, the advantage of disentangled representation on OOD generalization still requires further in-depth research and discussion.

\section{Supervised Model Learning for OOD Generalization}
\label{sec:causal}

Aside from the solely unsupervised learning of representations, there are a multitude of studies that incorporate supervised information (labels) to devise different model architectures and learning strategies. These methods place emphasis on end-to-end model learning to enhance their performance in OOD generalization. Given the vast range of literature in this field, we further categorize these methods based on their additional information requirements. 
This categorization facilitates a fairer and more comprehensible comparison of the various approaches.

\subsection{Require Environment Labels}
Numerous existing methods strive to exploit explicit environment labels to improve OOD generalization. 
Notably, these approaches encompass causal learning, invariant learning, and a variety of training strategies. 
In this section, we encapsulate the fundamental concepts underlying these techniques.

\paragraph{Causal Learning} Causal learning methods aim to learn the underlying causal structure of the data and to predict the outcome variable based on the identified causal variables. By correctly identifying the cause-effect relationships, these methods are expected to perform well even when the data distribution changes, as the underlying causal structure is often assumed to remain invariant across different environments or domains.

\paragraph{Invariant Learning} Invariant learning methods aim to learn features or representations that are invariant across different environments. The idea is that by focusing on the aspects of the data that do not change across environments, models could generalize better to new, unseen environments.

\paragraph{Training Strategies} Certain training strategies also utilize explicit environment labels to improve generalization. These methods often involve training models in a way that explicitly takes into account the potential differences between environments. For example, some methods might involve training separate models for each environment or explicitly modeling the differences between environments.

\subsubsection{Causal Learning}
Causal learning, rooted in the causal inference literature, provides a principled approach to the problem of OOD generalization. 
Its primary goal is to leverage causal variables for predictions, making it increasingly practical in recent times. 
We start with an introduction to the foundational concepts of causal learning, followed by an exploration of various related studies.

The underlying assumption of causal learning is captured in Assumption \ref{assumption: causality}, which originates in the causal inference literature. 
It postulates the existence of a causally invariant relationship between the target variable $Y$ and its direct causes $X_{\text{pa}(Y)}$.
This assumption implies that causal variables $X_{\text{pa}(Y)}$ remain stable across different environments or despite biases in data selection. 
This stability has driven a range of studies aimed at achieving OOD generalization through the exclusive exploitation of causal variables.

\begin{assumption}[Causality Assumption~\cite{buhlmann2018invariance}]
\label{assumption: causality}
	The structural equation models:
	\begin{align}
		Y^e &\leftarrow f_Y(X_{\text{pa}(Y)}^e, \epsilon_Y^e),\text{ $\epsilon_Y^e \perp X_{\text{pa}(Y)}^e$}
	\end{align}
	remains the same across all environments $e\in\text{supp}(\mathcal{E}_{all})$, that is, $\epsilon_Y^e$ has the same distribution as $\epsilon_Y$ for all environments. 
	$\textrm{pa}(Y)$ denotes the direct causes of $Y$.
\end{assumption}

Next, we explore methods related to causal inference, which endeavor to extract causal variables from heterogeneous data.
It's common knowledge that the gold standard for identifying the causal effect of a variable is to carry out randomized experiments, like A/B testing. 
However, these full-scale randomized experiments can be prohibitively expensive and often impractical in real-world applications. 
The ambitious nature of causal inference or causal structure learning makes these techniques more of an ideal "ground truth" rather than a practically achievable goal in typical machine learning settings. 

Therefore, it is more pragmatic to design techniques that provide a more "causal explanation" compared to the standard regression or classification framework, while also offering a degree of invariance across environments. Following this intuition, a series of methods \cite{peters2016causal, Pfister2018, rothenhausler2018anchor, Heinze-Deml2018, Gamella2020, Oberst2021} have been proposed, leveraging the inherent heterogeneity within data (e.g., across multiple environments).

\begin{assumption}[Invariance Assumption] 
\label{assumption:invariance}
	There exists a subset $S^*\subseteq \{1,\ldots ,p\}$ of the covariate indices (including the empty set) such that  
	\begin{eqnarray}
		P(Y^e|X_{S^*}^e)\ \mbox{is the same for all}\ e \in {\cal E}.
	\end{eqnarray}
	That is, when conditioning on the covariates from $S^*$ (denoted by
	$X^e_{S^*}$), the conditional distribution is invariant across all environments from ${\cal E}$.
\end{assumption}

\citet{peters2016causal} first try to investigate the fact that "invariance" could, to some extent, infer the causal structure under necessary conditions and propose Invariant Causal Prediction (ICP).
Specifically, they leverage the fact that when considering all direct causes of a target variable, the conditional distribution of the target given the direct causes will not change when interfering all other variables in the model except the target itself.
Then they perform a statistical test whether a subset of covariates $S$ satisfies the invariance assumption \ref{assumption:invariance} for the observed
environments in $\mathcal{E}$.
The null hypothesis for testing is: 
\begin{eqnarray*}
	H_{0,S}({\cal E}):\ \mbox{invariance assumption}\ \mbox{holds}.
\end{eqnarray*}
and all subsets of covariates $S$ which lead to invariance are intersected, that is: 
\begin{eqnarray*}\label{icp}
	\hat{\mathcal S}({\mathcal E}) = \bigcap_{S} \{S;\ H_{0,S}({\mathcal E})\ \mbox{not rejected by test at significance level}\ \alpha\}. 
\end{eqnarray*}
Under the assumption of structural equation model and Gaussian residual described in \cite{peters2016causal}, ICP with Chow test \cite{chow1960tests} could, at least with controllable probability 1-$\alpha$, discover subsets of true causal variables, which reads as:
\begin{eqnarray}
	\mathbb{P}[\hat{\mathcal S}({\mathcal E}) \subseteq \mathrm{pa}(Y)] \ge 1- \alpha,
\end{eqnarray}
where $\mathrm{pa}(Y)$ denotes the direct causes of target $Y$ (e.g. the parental variables of $Y$ in the causal graph).
Though being the first attempt to connect invariance with causality, ICP has several limitations.
The most straightforward one is the strict requirements for heterogeneity since the power of ICP depends highly on the quality of available environments $\mathcal{E}_{tr}$(or perturbations).
If the available perturbed subpopulations are not enough, or even a single environment, the efficacy of ICP will be lost.
As discussed in \citep{Pfister2018}, naively estimating the environments from data and then applying ICP may yield less powerful results, so instead of using static data, \citet{Pfister2018} propose to leverage the sequential data from a non-stationary environment to detect instantaneous causal relations in multivariate linear time series, which relaxes the assumption of known environments.
Besides environmental specification, there are other works trying to consolidate the coverage of the so-called invariance-based method.
For example, \citet{Heinze-Deml2018} extend the ICP into non-linear model and continuous environments; \citet{Gamella2020} apply the ICP into an active learning setting where the interventions (a.k.a. environments) can be proactively chosen during training.

\begin{figure}[htbp]
	\centering
	\subfloat[Traditional IV model.]{
	\label{subfig:anchor-a}
		\includegraphics[width=0.25\linewidth]{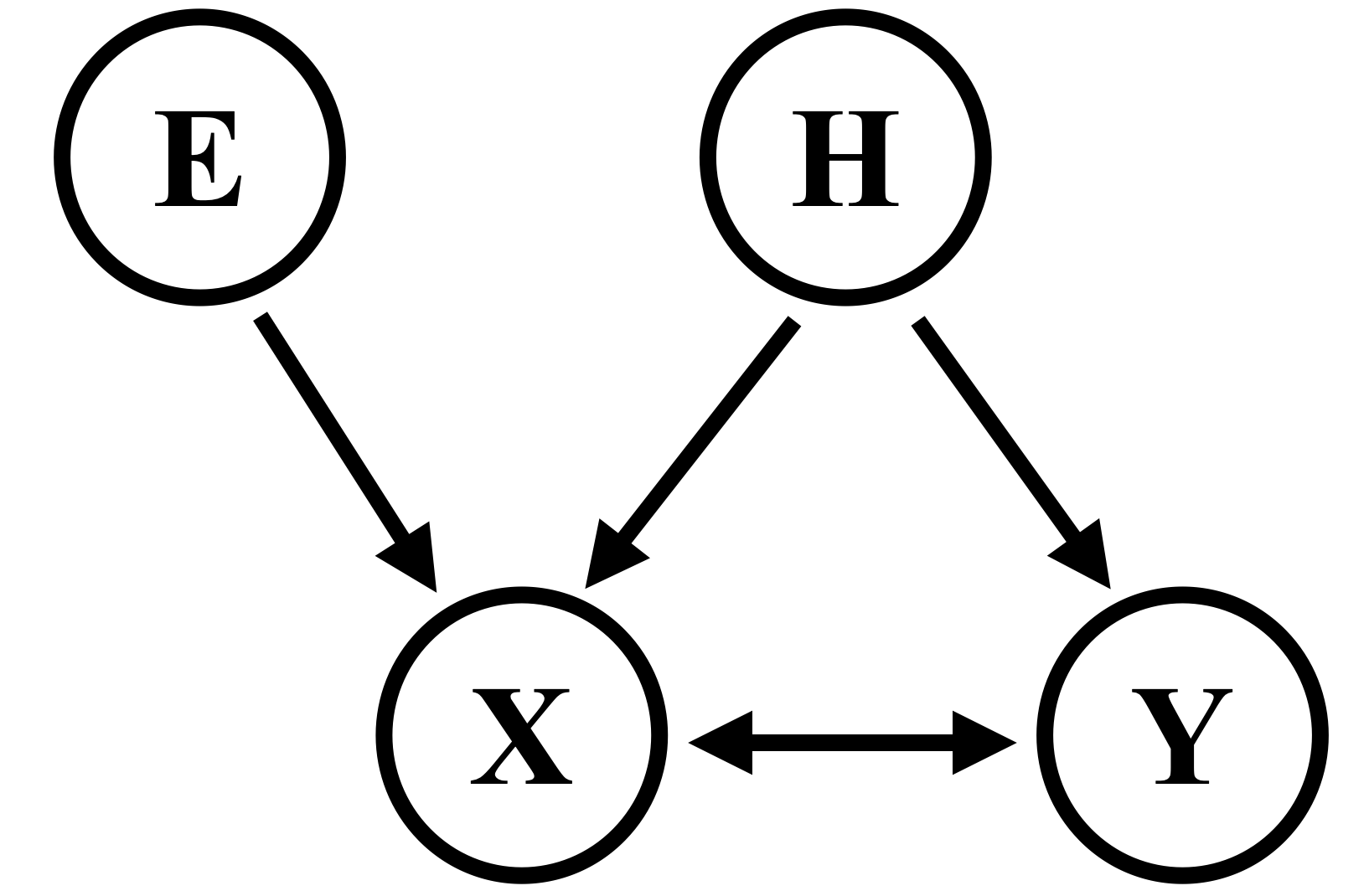}
	}
	\hspace{0.15in}
	\subfloat[Anchor regression model.]{
	\label{subfig:anchor-b}
		\includegraphics[width=0.25\linewidth]{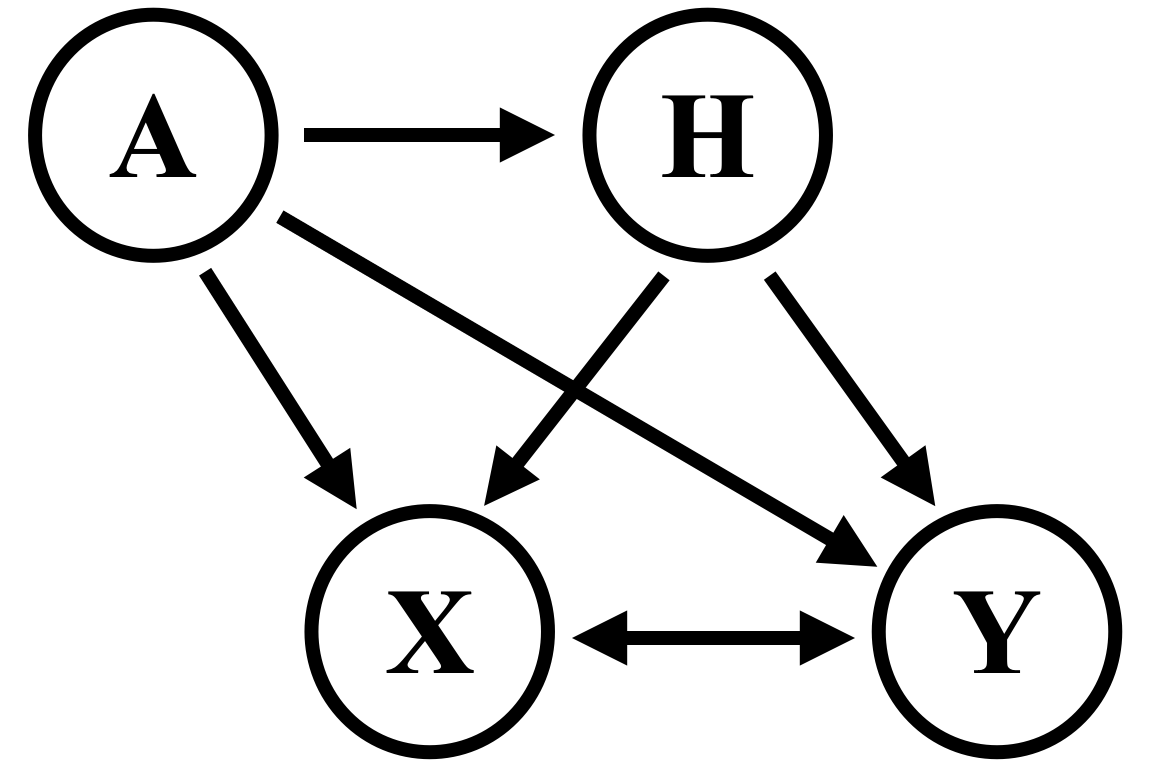}
	}
	\caption{Comparision of SCM between IV model and Anchor Regression model.}
	\label{fig:anchor}
\end{figure}

ICP serves as a milestone towards inferring causal structure via invariance property.
However, the invariance assumption may be violated in more complicated scenarios.
Among which, the most common case is the existence of hidden confounders.
The instrument variable(IV) method is one typical method for dealing with hidden confounders, which requires the instrument variable $E$ not to directly act on the hidden confounding variable $H$ and outcome variable $Y$, as shown in Figure \ref{subfig:anchor-a}.

\citet{rothenhausler2018anchor} investigate more relaxed conditions than the standard IV model which allow the direct effect of instrument variable (which they called anchor variables) on $H$ and $Y$, as shown in Figure \ref{subfig:anchor-b}.
They realize that despite the attractive notion of invariance guarantee against arbitrarily large intervention or perturbation, one seldom encounters such extreme cases, and exact invariance could be too conservative for moderately perturbed data.
Specifically, they focus on the following structural equation:
\begin{eqnarray}
	Y = X^T \beta + H^T \alpha + A^T \xi + \epsilon_Y,
\end{eqnarray}
with $X \in \R^p,\ H \in \R^q$ and $A \in \R^r$, and proposed a regularized formulation of ordinary least squares model by the error projection to the space spanned by anchor variables.
\begin{eqnarray}\label{anchor-est}
\hat{\beta}(\gamma) = \arg\min_b \left(\|(I - \Pi_{\mathbf{A}})(\textbf{Y} - \textbf{X}
		b)\|_2^2/n + \gamma \|\Pi_{\mathbf{A}}(\textbf{Y} - \textbf{X} b)\|_2^2/n \right).
\end{eqnarray}
where $\Pi_{\mathbf{A}}$ denotes the projection in $\R^n$ onto the column space of $\mathbf{A}$. 
For $\gamma = 1$, $\hat{\beta}(1)$ equals the ordinary least squares estimator, for $\gamma \to \infty$ it obtains the two-stage least squares procedure from IV regression.

When the observation of instrument $A$ is still hard to fulfill, \citet{Oberst2021} further relax the assumption by introducing the noisy proxy of $A$ and prove the robustness of the method under bounded shifts.
Based on these, \citet{pmlr-v216-mazaheri23a} focus on the setting where the causal and anti-causal variables of the outcome variable are unobserved and propose feature selection and engineering methods to identify proxies.

\subsubsection{Invariant Learning}
\label{subsubsec:Rep-DG}
Relaxing the causality to invariance, the key idea of invariant learning is to learn an invariant representation or model across environments leveraging contextual information such as domain labels.   
There are works~\cite{muandet2013domain, arjovsky2019invariant,albuquerque2020adversarial} that theoretically or empirically prove that if the representations remain invariant when the domain varies, the representations are transferable and robust on different domains.  
Methods that seek invariance among different environments can be mainly divided into three categories, namely invariant risk minimization and domain-irrelevant representation learning.

\paragraph{Invariant Risk Minimization}
\label{subsubsec:inv}
Deriving from causal inference, invariant risk minimization (IRM, \citep{arjovsky2019invariant}), targets on latent causal mechanisms and extends ICP to more practical and general settings.  
Different from causal learning methods that act on raw variable level, IRM makes the invariance assumption as:
\begin{assumption}[IRM's Invariance Assumption]
\label{assumption: IRM-invariance}
	There exists data representation $\Phi(X)$ such that for all $e,e' \in \text{supp}(\mathcal{E}_{tr})$, $\mathbb{E}[Y|\Phi(X^e)]=\mathbb{E}[Y|\Phi(X^{e'})]$, where $\mathcal{E}_{tr}$ denotes the available training environments.
\end{assumption} 
\citet{arjovsky2019invariant} propose to find data representation $\Phi(X)$ that can both predict well and elicit an invariant linear predictor $w$ across $\mathcal{E}_{tr}$, which results in the following objective function:
\begin{align}
\label{equ:IRM}
	&\ \ \min\limits_{\Phi(X), w}\sum_{e\in\text{supp}(\mathcal{E}_{tr})} \mathcal{L}^e(w\odot \Phi(X),Y)\\
	&\text{s.t. } w\in \arg\min_{\overline{w}}\mathcal{L}^e(\overline{w}\odot\Phi(X)),\text{ for all }e\in\text{supp}(\mathcal{E}_{tr})
\end{align}
In order to achieve invariance across $\mathcal{E}_{all}$ by enforcing low error of \Cref{equ:IRM} on $\mathcal{E}_{tr}$, IRM requires sufficient diversity across environments and makes the following assumption.

\begin{assumption}[IRM's Condition, Assumption 8 in \cite{arjovsky2019invariant}]
\label{assumption:IRM2}
	A set of training environments $\mathcal{E}_{tr}$ lie in linear general position of degree $r$ if $|\mathcal{E}_{tr}|>d-r+\frac{d}{r}$ for some $r\in \mathbb{N}$, and for all non-zero $x\in\mathbb{R}^d$:
	\begin{small}
	\begin{equation}
		\mathrm{dim}\left(\mathrm{span}\left(\{\mathbb{E}_{X^e}[X^e{X^e}^T]x - \mathbb{E}_{X^e, \epsilon^e}[X^e\epsilon^e]\}_{e\in\mathcal{E}_{tr}}\right)\right) > d-r
	\end{equation} 
	\end{small}
\end{assumption} 

With Assumption \ref{assumption:IRM2}, IRM characterizes under what conditions an invariant predictor $w\circ \Phi(X)$ across $\mathcal{E}_{tr}$ remains invariant across $\mathcal{E}_{all}$ in linear cases (Therorem 9 in \cite{arjovsky2019invariant}).
It makes the same assumptions about linearity, centered noise, and independence between the noise $\epsilon^e$ and the causal variables as ICP~\citep{peters2016causal}, but does not assume that the data is Gaussian, the existence of a causal graph, or that the training environments arise from specific types of interventions.
And the result of IRM could extend to latent causal variables while ICP \citep{peters2016causal} restricts to raw causal feature level.

Based on IRM, follow-up works have proposed variations on this objective with similar regularizations of the invariance assumption \ref{assumption: IRM-invariance}, resulting in similar alternatives.
\citet{DBLP:rationalization} and \citet{DBLP:journals/corr/abs-2008-01883} formulate the desired invariant representation using information theory, and propose to find the maximal invariant predictor (MIP) across training environments.
The maximal invariant predictor is defined as:
\begin{definition}
	\label{definition2:invariance set}
	The invariance set $\mathcal{I}$ with respect to $\mathcal{E}$ is defined as:
		\begin{equation}
			\begin{aligned}
				\mathcal{I}_{\mathcal{E}} = \{\Phi(X): Y \perp \mathcal{E}|\Phi(X)\}
				= \{\Phi(X): H[Y|\Phi(X)]=H[Y|\Phi(X),\mathcal{E}]\}
			\end{aligned}
		\end{equation}
	where $H[\cdot]$ is the Shannon entropy of a random variable. 
	The corresponding maximal invariant predictor (MIP) of $\mathcal{I}_{\mathcal{E}}$ is defined as:
	\begin{equation}
		S_{\mathcal{E}} = \arg\max_{\Phi \in \mathcal{I}_{\mathcal{E}}}I(Y;\Phi)
	\end{equation}
	where $I(\cdot;\cdot)$ measures Shannon mutual information between two random variables.
\end{definition}
With the invariant predictor $S_{\mathcal{E}_{all}}$, \citet{DBLP:journals/corr/abs-2008-01883} prove that the OOD optimal model is given by $\mathbb{E}[Y|S_{\mathcal{E}_{all}}]$.
Further, to obtain the MIP solution from training environments $\mathcal{E}_{tr}$, \citet{DBLP:journals/corr/abs-2008-01883} derive a regularizer as:
\begin{equation}
	\text{Trace}\left(\mathrm{Var}_{e\in\mathcal{E}_{tr}}(\nabla_\theta \mathcal{L}^e(\theta))\right)
\end{equation}
where the variance is taken with respect to training environments $\mathcal{E}_{tr}$.
Under the controllability condition proposed by \citet{DBLP:journals/corr/abs-2008-01883} which assumes that there exists an environment $e$ such that $X\perp Y | \Phi(X),e$, the optimality of $\mathbb{E}[Y|\Phi(X)]$ can be verified, as shown in Proposition 3.1 in \citep{DBLP:journals/corr/abs-2008-01883} and Theorem 1 in \citep{rojas2018invariant}.

Moreover, \citet{ahuja2020invariant} introduce game theory to this field and substitute the linear classifier in IRM with an ensemble of classifiers originating from different environments.
\citet{jin2020domain} replace IRM's regularizer with predictive regret, imposing more stringent constraints on $\Phi(X)$.
\citet{krueger2021out} propose penalizing the variance of risks across different environments, while \citet{xie2020risk} suggest a similar objective but swap the original penalty with the square root of the variance.
\citet{pmlr-v139-mahajan21b} present a contrastive regularizer that matches the representation of identical objects across different environments.
\citet{creager2020environment} focus on IRM's issue of missing environment labels and put forward Environment Inference for Invariant Learning (EIIL) to maximize IRM's penalty by learning environments. This two-stage algorithm first generates environments based on a biased reference model, then carries out invariant learning using the environments learned.
\citet{Xin_Wang_Su_Wang_2023} elucidate the link between invariant learning and adversarial training for OOD generalization and introduce an adversarial training method to mitigate distribution shifts.
\citet{fan2023environment} examine the invariant learning problem from a statistical perspective and propose an environment-invariant linear least squares objective function.
And \citet{Li_Shen_Wang_Zhu_Reed_Li_Keutzer_Zhao_2022} suggest an invariant information bottleneck to extract invariant representations via mutual information.

While the results from IRM seem encouraging, \citet{rosenfeld2020risks} highlight some issues with its application to classification tasks. 
In the linear case, they offer simple conditions under which the optimal solution either succeeds or, more often, fails to recover the optimal invariant predictor. 
Notably, \citet{rosenfeld2020risks} prove that a viable solution can outperform the optimal invariant predictor on all $e\in\mathcal{E}_{all}$, using only environmental features (Theorem 5.3 in \cite{rosenfeld2020risks}). 
In a nonlinear context, they illustrate that IRM can fail dramatically unless the test data closely resemble the training distribution (Theorem 6.1 in \cite{rosenfeld2020risks}).
Furthermore, \citet{kamath2021does} demonstrate that it's possible for IRM to learn a sub-optimal predictor, due to non-invariant loss function across environments.
\citet{ahuja2020empirical} compare IRM with ERM from a sample complexity perspective across different shift patterns (Table 1 in \citep{ahuja2020empirical}). 
They conclude that under covariate shifts, IRM doesn't show a clear advantage over ERM. However, in the case of other distribution shifts involving confounders or anti-causal variables, IRM is likely to be close to the desired OOD solutions within the finite sample regime. 
Recently, drawing inspiration from PAC learning, \citet{pmlr-v202-parulekar23a} provide finite-sample OOD generalization guarantees for approximate invariance. Meanwhile, \citet{pmlr-v162-wang22x} manage to reduce the required number of training environments to $\mathcal O(1)$ by using second-order moment information.

\paragraph{Domain-irrelevant representation learning}

\citet{ganin2015unsupervised, ganin2016domain} first introduce the concept of the Domain-Adversarial Neural Network (DANN) for domain adaptation. The goal of DANN is to cultivate representations that are both discriminative and impervious to domain shifts. This is accomplished by jointly optimizing the base features and a label predictor that predicts class labels during both training and inference phases. Concurrently, a domain classifier is trained to distinguish between source and target domains. By developing representations that confuse this domain classifier, features invariant to the domain are cultivated.
Building on this, \citet{li2018domain} adapt this framework for scenarios where the target domain information is unknown, while \citet{gong2019dlow} extend adversarial training into the manifold space. \citet{li2018deep} suggest the utilization of class-specific adversarial networks through a Conditional Invariant Adversarial Network (CIAN). 
Providing a theoretical foundation, \citet{garg2021learn} and \citet{sicilia2021domain} derive generalization bounds for domain adversarial training. \citet{rahman2020correlation} propose a correlation-aware adversarial framework that can be applied to both Domain Adaptation (DA) and OOD generalization. This framework leverages both the correlation alignment metric and adversarial learning to minimize the domain discrepancy between the source and target data.
On the application front, \citet{shao2019multi}, \citet{jia2020single}, and \citet{wang2020unseen} apply domain adversarial learning for face anti-spoofing and unseen target stance detection. Moreover, \citet{zhao2020domain} introduce an entropy-regularization approach that learns invariant representations by minimizing the KL-divergence between the conditional distributions of different source domains.

In addition to domain adversarial learning, numerous studies~\cite{tzeng2014deep, li2018domain, long2014transfer, long2013transfer, long2013adaptation, xu2021fourier} propose the alignment of features to cultivate domain invariant representations. 
\citet{motiian2017unified} recommend semantic alignment between various domains, achieved by minimizing the distance between samples of the same class but different domains, and maximizing the distance between samples from different classes and domains. 
Other research focuses on minimizing feature distribution divergence by reducing the Maximum Mean Discrepancy (MMD) distance~\cite{pan2010domain, tzeng2014deep, wang2018visual}, Wasserstein distance~\cite{zhou2020domain}, or second-order correlation~\cite{sun2016deep, sun2016return, peng2018synthetic}, for either Domain Adaptation (DA) or Out-of-Distribution (OOD) generalization.
In more recent work, \citet{10093034} integrate knowledge distillation to learn both domain-invariant and domain-specific representations. \citet{Yu_2023_CVPR} employ a diffusion model to align training and testing distributions reversibly. 
\citet{Lv_2022_CVPR} suggest extracting causal factors from input data and then reconstructing the invariant causal mechanisms. \citet{Wang_2022_CVPR} implement causal invariant transformations that disturb only non-causal features to achieve invariance. 
\citet{NEURIPS2022_82e330c1} exploit the shared causal structure of domains to learn invariant and transferable representations. Meanwhile, \citet{DBLP:conf/emnlp/JiaZ22} learn distributional invariance across source domains to align vocabulary and feature distributions using prompting.

\paragraph{Applications}
Nowadays, invariant learning has found wide-ranging applications, often yielding improved generalization performance. 
\citet{DBLP:conf/iclr/WuWZ0C22}, \citet{NEURIPS2022_8b21a7ea}, and \citet{gui2023joint} have applied invariant learning to graph data, achieving enhanced out-of-distribution (OOD) generalization performance.
In the field of drug discovery, \citet{NEURIPS2022_54710808} propose to learn sub-structure invariance for OOD molecular representations.
Expanding the concept of invariance into the realm of reinforcement learning, \citet{10005169} introduce the notion of policy invariance. 
To mitigate the effects of unobserved confounders in recommendation systems, \citet{10.1145/3534678.3539439} apply invariant learning.
And \citet{10.1145/3503161.3548035} introduce invariant grounding to foster interpretability in video question-answering tasks.

\subsubsection{Training Strategy}
\label{subsubsec:Tr-DG}
In the field of Out-of-Distribution (OOD) generalization, several vision-related studies have focused on the development of training strategies designed to enhance the generalization capability of deep learning models applied to image data. 
These strategies can be broadly categorized into five key areas: Meta-Learning, Ensemble Learning, Self-Supervised Learning, Feature Normalization, and Prompt Tuning.

\paragraph{Meta-learning}
Meta-learning establishes a unique learning paradigm, wherein knowledge is accrued over multiple learning episodes~\cite{hospedales2020meta}. The concept of "episodes" in the training phase is first introduced by \citet{finn2017model} in their model-agnostic meta-learning (MAML) approach for Domain Adaptation (DA). 
This concept has since greatly influenced research in the field of meta-learning for OOD generalization. 
In the seminal work of \citet{li2018learning}, meta-learning is first applied to OOD generalization, paving the way for numerous subsequent improvements~\cite{balaji2018metareg, otalora2019staining, li2019feature, dou2019domain,liu2020shape, du2020learning, du2020metanorm,zhao2021learning, choi2021meta}. 
The primary approach involves dividing the source domains into meta-train and meta-test sets. The model is then trained to simultaneously optimize both the meta-train and meta-test losses.

\paragraph{Model-ensemble learning.}
Model-ensemble learning methods typically aim to improve generalization capability by utilizing an ensemble of distinct models, each one tailored for different source domains. 
Several studies adopt domain-specific subnetworks corresponding to individual source domains, while employing a singular classifier~\cite{zhou2020domain, d2018domain,ding2017deep} or multiple domain-specific classifier heads~\cite{wang2020dofe}. 
Alternatively, some methods use domain-specific batch normalization for different domains to achieve better normalization results~\cite{segu2020batch, mancini2018best}.
Recently, given the promising performance of pre-trained models, a study by \citet{DBLP:conf/nips/DongAZXHYBL22} evaluate the inter-class discriminability and inter-domain stability of these models, and construct an ensemble of top-ranked models. This approach achieves state-of-the-art performance on the DomainNet dataset \citep{DBLP:conf/iccv/PengBXHSW19}.

\paragraph{Self-supervised learning}
Drawing inspiration from self-supervised learning methods~\cite{noroozi2016unsupervised}, \citet{carlucci2019domain} combine a self-learning puzzles task with the classification task to learn robust representations. 
\citet{ryu2019generalized} devise a strategy to sample positive and negative instances using a random forest. 
In an alternating training approach, \citet{li2019episodic} separately train the convolutional layers and the classifier. 
To guard against overfitting to source domains, \citet{huang2020self} introduce a self-challenging dropout algorithm.

\paragraph{Feature Normalization} 
Several studies~\cite{sun2016deep, sun2016return, jin2020domain} have employed feature normalization to mitigate domain discrepancy.
\citet{pan2018two} introduce IBN-Net, which ingeniously integrates Instance Normalization (IN) \cite{ulyanov2017improved} and Batch Normalization (BN)~\cite{ioffe2015batch} as building blocks to both capture and reduce domain variance. 
Empirically, they discover that while IN provides visual and appearance invariance, it could potentially diminish the discriminative information of representations. 
Consequently, they suggest a combination of IN and BN in shallow layers, with only BN being employed in deeper layers. 
In the realm of style transfer, \citet{huang2017arbitrary} propose the utilization of adaptive IN. 
Following the assumption that each feature map of a convolutional encoder can be segregated into style-related and shape-related components, \citet{nam2018batch} explicitly combine BN and IN to learn style-invariant representations. 
Lastly, \citet{jin2021style} present Style Normalization and Restitution (SNR), a method that distills task-related features from the residual post style normalization, thereby ensuring the discriminability of the features.

\paragraph{Prompt tuning}
Recently, with the remarkable zero-shot generalization exhibited by pre-trained vision-language models across a variety of downstream tasks, several works have sought to leverage prompt tuning during testing to further enhance the model's ability to generalize to unseen domains. 
\citet{DBLP:conf/nips/ShuNHYGAX22} introduce an approach to learn adaptive prompts on the fly using a single test sample. 
This demonstrates superior generalization performance when compared to prior prompt tuning methodologies. 
In a different approach, \citet{DBLP:journals/corr/abs-2306-15955} optimize prompts employing equiangular tight frame (ETF) structures. 
\citet{DBLP:journals/corr/abs-2208-08914} integrate domain prompts into the Vision Transformer architecture to improve its generalization capabilities.

\subsection{No Environment Labels}
In this section, we present methods that do not necessitate explicit environment labels.
These approaches include stable learning, heterogeneity-aware learning, and others like flatness-aware learning.

\subsubsection{Stable Learning}
\label{subsec:stable}
Stable learning, as compared to causal and invariant learning, offers an alternative method for integrating causal inference with machine learning, significantly reducing the dependency on environment labels. The problem setting for stable learning is as follows:
\begin{problem}[Settings of Stable Learning]
Given training data $D^{e}=(X^{e},Y^{e})$ from \textbf{one} environment $e\in \text{supp}(\mathcal{E}_{all})$, the goal of stable learning is to learn a predictive model with \textbf{uniformly good performance} on any possible environments in $\text{supp}(\mathcal{E}_{all})$. 
\end{problem}

To address this problem, drawing inspiration from variable balancing strategies as seen in the literature~\cite{athey2016approximate,zubizarreta2015stable,hainmueller2011entropy}, \citet{shen2018causally} propose treating all variables as potential treatments and learning a set of global sample weights. 
These weights serve to remove any confounding bias for all potential treatments from the data distribution. 
They develop a global balancing loss that can be seamlessly integrated as a regularizer into standard machine learning tasks, as illustrated by \Cref{equ:causal_regularizer}:
\begin{equation}
\label{equ:causal_regularizer}
\sum_{j=1}^{p} \left\|\frac{\bm{X}_{-j}^{T}\cdot(W\odot \bm{I}_j)}{W^{T}\cdot \bm{I}_j}-\frac{\bm{X}_{-j}^{T}\cdot(W\odot (1-\bm{I}_j))}{W^{T}\cdot (1-\bm{I}_j)}\right\|_2^2,
\end{equation}
where $W$ represents the sample weights, $\big|\frac{\bm{X}_{-j}^{T}\cdot(W\odot \bm{I}_j)}{W^{T}\cdot \bm{I}j}-\frac{\bm{X}{-j}^{T}\cdot(W\odot (1-\bm{I}_j))}{W^{T}\cdot (1-\bm{I}_j)}\big|2^2$ signifies the loss of confounder balancing when setting feature $j$ as the treatment variable. 
Here, $\bm{X}{-j}$ refers to all remaining features (i.e., confounders) excluding the $j^{th}$ column. 
$\bm{I}j$ stands for the $j^{th}$ column of $\bm{I}$, and $\bm{I}{ij}$ indicates the treatment status of unit $i$ when feature $j$ is treated as the treatment variable. 
By minimizing the global balancing loss, it's possible to remove the confounding bias on a global scale.
Furthermore, \citet{kuang2018stable} integrate unsupervised feature representation into the global balancing stage using auto-encoders~\cite{Sch2007Greedy} and adapt the original regularizer into a "deep" version.

The aforementioned methods primarily focus on binary features as mainstream discussions on causal inference predominantly involve binary treatments. 
However, when the treatment variable is categorical or continuous, traditional balancing methods become infeasible due to the potentially infinite treatment levels. 
To address this limitation, especially in situations involving continuous treatments, \citet{kuang2020stable} propose a solution to learn a set of sample weights. 
These weights are tailored such that the weighted distribution of the treatment and confounder meet the condition of independence. 
This corresponds to the fact that accurate treatment effect estimates can be obtained if the treatment and confounder are independent.

In addition to methods addressing confounder bias, \citet{shen2020stable} focus on the issue of model misspecification for linear models within the context of stable learning. 
The primary challenge for stable learning in linear cases stems from the unavoidable model misspecification that typically occurs in real-world scenarios. 
More specifically, the true generative process often contains not just the linear part, but also an additional misspecification term. 
This term could be a nonlinear element or interactions between input variables.
\begin{equation}
y=x^{\top} \bar{\beta}_{1 ; p}+\bar{\beta}_{0}+b(x)+\epsilon.
\end{equation}
\citet{shen2020stable} reveal that the collinearity between variables is a crucial factor in achieving a stable model. 
If a mis-specified model is used at the training phase, the presence of collinearity amongst variables can escalate a minor mis-specification error to an arbitrarily large magnitude, resulting in unstable prediction performance across variably distributed test data.
To mitigate the effects of collinearity among variables, \citet{shen2020stable} propose to learn a set of sample weights that promote near orthogonality in the design matrix. Technically, they construct an uncorrelated design matrix, denoted as $\tilde{X}$, from the original $X$ matrix, treating it as the 'oracle'. They then learn the sample weights $w(x)$ by estimating the density ratio $w(x)=p_{\tilde{D}}(x)/p_{D}(x)$ between the underlying uncorrelated distribution $\tilde{D}$ and the original distribution $D$.

To further mitigate the issues of large variance and the shrinkage of the effective sample size introduced by sample reweighting, \citet{shen2020stableD} suggest leveraging unlabeled data gathered from multiple environments to uncover hidden cluster structures among variables. Under several technical assumptions, they demonstrate that decorrelating variables between clusters, rather than among themselves, is sufficient for achieving stable estimation without inflating the variance.
In contrast, \citet{DBLP:journals/corr/abs-2212-00992} propose an iterative framework that combines sample reweighting and a sparsity constraint to alleviate these issues, even without access to multiple environments. They provide theoretical proof that the introduction of a sparsity constraint can help lessen the requirement for large sample sizes when selecting stable variables.

Recently, \citet{zhang2021deep} propose StableNet, which extends former linear frameworks~\cite{kuang2018stable, shen2020stable, kuang2020stable} to incorporate deep models.
Due to the complexity of nonlinear dependencies among features derived from deep models, it is significantly more challenging to measure and eliminate the dependencies among features compared to linear cases. In response, StableNet introduces an innovative approach to nonlinear feature decorrelation, leveraging Random Fourier Features (RFF)~\cite{rahimi2007random}.
Specifically, StableNet iteratively optimizes sample weights $\mathbf{w}$, representation function $f$, and prediction function $g$ as follows:
\begin{equation} \label{eq:overall}
    \begin{aligned}
        f^{(t+1)}, g^{(t+1)} = & \underset{f, g}{\arg \min} \sum_{i=1}^n w^{(t)}_i \mathcal{L}(g(f(\mathbf{X}_i)), y_i), \\
        \mathbf{w}^{(t+1)} = & \underset{\mathbf{w} \in \Delta_n}{\arg \min} \sum_{1 \le i < j \le m_Z} \left\Vert\hat{\Sigma}_{\mathbf{Z}^{(t+1)}_{:,i}\mathbf{Z}^{(t+1)}_{:, j};\mathbf{w}}\right\Vert_F^2. \\
    \end{aligned}
\end{equation}
where $\mathbf{Z}^{(t+1)}=f^{(t+1)}(\mathbf{X})$, $\mathcal{L}(\cdot, \cdot)$ represents the cross entropy loss function and $t$ represents the time stamp. 
In StableNet, the sample reweighting module and the representation learning network are jointly optimized. This efficient cooperation facilitates the isolation of environment-related features, thereby utilizing truly category-related and discriminative features for prediction. Consequently, StableNet can deliver more stable performances in non-stationary environments in the wild.

Subsequent studies building upon StableNet~\cite{zhang2021deep} have extended the feature decorrelation-based reweighting techniques to the areas of graph data~\cite{fan2021generalizing} and natural language understanding~\cite{dou2022decorrelate}. 
Additionally, \citet{zhang2022towards} has broadened the scope of the OOD problem to object detection, examining the impact of distribution shifts in that field.
In another extension, \citet{DBLP:journals/corr/abs-2006-04381} adopt subsampling techniques to mitigate the confounding effects brought about by distributional shifts. Also, by integrating the decorrelation mechanism, \citet{DBLP:conf/ijcai/WangFKSLW20} is able to enhance clustering performance even under data selection bias.
Further, \citet{DBLP:conf/mm/ZhangJWKZZYYW20} propose a Deconfounded Visio-Linguistic Bert framework aimed at curbing potential data biases. 
Meanwhile, \citet{9428205} introduce a method for identifying causal features using a meta-learning mechanism for OOD generalization.

\subsubsection{Heterogeneity-Aware Learning}
\label{subsec:Inv-opt}
In addition to stable learning methods, several strategies have been developed to leverage the latent heterogeneity within data to enhance OOD generalization capabilities. 
In realistic scenarios, data often originates from various sources, often without explicit environmental labels, which makes multiple environments inaccessible.
Moreover, it may be challenging to pre-define the types of environments needed, particularly in complex real-world applications such as recommendation systems, where numerous kinds of biases are present. 
Furthermore, theoretical analysis conducted in \cite{hrm} shows that achieving desired invariance properties becomes particularly challenging when environments are not appropriately characterized.

In order to mitigate these practical issues, several works \cite{hrm,KerHRM,creager2020environment,liu2022distributionally,DBLP:conf/iclr/LiuWPXZ0023, DBLP:journals/corr/abs-2304-00305} have sought to uncover and utilize the latent heterogeneity within data to improve invariance and generalization capabilities. \citet{DBLP:conf/iclr/LiuWPXZ0023} present the first quantitative definition of predictive heterogeneity using $\mathcal V$-information \cite{DBLP:conf/iclr/XuZSSE20}, and propose an information maximization algorithm to explore this latent heterogeneity.
\citet{hrm} introduce Heterogeneous Risk Minimization (HRM), an optimization framework that simultaneously learns the heterogeneity within data and the invariant predictor. This framework consists of two interactive components: a frontend module, $\mathcal{M}_c$, for heterogeneity identification, and a backend module, $\mathcal{M}_p$, for invariant prediction. 
Given the pooled heterogeneous data, the process begins with the heterogeneity identification module $\mathcal{M}_c$, which uses the learned variant representation $\Psi(X)$ to generate heterogeneous environments $\mathcal{E}_{learn}$. These learned environments are then utilized by the OOD prediction module $\mathcal{M}_p$ to learn the Maximal Invariant Predictor (MIP~\citep{DBLP:journals/corr/abs-2008-01883}) $\Phi(X)$, as well as the invariant prediction model $f(\Phi(X))$. Subsequently, a better variant $\Psi(X)$ is derived from the learned MIP $\Phi(X)$, which further enhances the heterogeneity identification process.
\citet{KerHRM} extend HRM to handle more complex data using the neural tangent kernel (NTK~\citep{NEURIPS2018_5a4be1fa}).
Recently, \citet{liu2022distributionally} provided a theoretical analysis of the invariant learning problem under latent heterogeneity, introducing the $\alpha_0$-distributional invariance property as a relaxation of the strict invariance property. They analyze the learnability and the generalization gap bound for an OOD generalization problem within the context of this new property. In addition, \citet{pmlr-v139-liu21f} found that training the model twice, with higher sample weights for harder samples, yields good generalization performance. Furthermore, \citet{pmlr-v177-idrissi22a} propose a series of straightforward data balancing methods that achieve competitive performances for the worst-group.

\subsubsection{Other Emergent Directions}
\paragraph{Flatness-aware learning}
Recently, the field of flatness-aware learning in deep neural networks has garnered significant attention, delivering state-of-the-art results on several image datasets. \citet{DBLP:conf/nips/ChaCLCPLP21} proposed Stochastic Weight Averaging Densely (SWAD) as a means of identifying flat minima solutions, which have demonstrated strong OOD generalization performance across five domain generalization benchmarks.

Furthering this research, \citet{FAD} contest that Adam~\cite{DBLP:journals/corr/KingmaB14}, despite being a popular choice, might not be the optimal optimizer for a large number of current OOD generalization methods. 
In response, they proposed a flatness-aware optimizer designed to efficiently locate both zeroth-order and first-order flat minima solutions. This optimizer has shown superior performance across a range of domain generalization datasets and benchmarks.

\section{Optimization for OOD Generalization}
\label{sec:optimization}
To address the Out-of-Distribution (OOD) generalization problem, apart from unsupervised representation learning and supervised model learning, robust optimization methods with theoretical guarantees have recently aroused much attention, which is both model agnostic and data structure agnostic, and therefore could be incorporated with various approaches.
In this section, we mainly focus on the literature of distributionally robust optimization (DRO), which stems from the literature on operations research and raises more and more attention from the machine learning community.

We first introduce the objective of these OOD optimization methods and then classify the methods according to the requirements of additional information.
In order to address the problem from the optimization perspective, the OOD generalization problem is formulated as the worst-case prediction error among $\mathcal{E}_{all}$, which takes the form of: 
\begin{equation}
	\label{equ:OOD}
	\arg\min_f \max_{e\in\text{supp}(\mathcal{E}_{all})} \mathcal{L}(f|e) 
\end{equation}
where $\mathcal{E}_{all}$ is the random variable on indices of all possible environments, and for all $e\in \mathrm{supp}(\mathcal{E}_{all})$, the data and label distribution $P^e(X,Y)$ can be quite different from that of training distribution $P_{tr}(X,Y)$; $\mathcal{L}(f|e)=\mathbb{E}[l(f(X),Y)|e]=\mathbb{E}^e[l(f(X^e),Y^e)]$ is the risk of predictor $f$ on environment $e$, and $l(\cdot,\cdot):\mathcal{Y}\times\mathcal{Y}\rightarrow\mathbb{R}^+$ is the loss function.
Intuitively, optimization methods aim to guarantee the worst-case performance under distributional shifts.

\subsection{No Additional Information}
\label{subsec:DRO}
Different from the aforementioned methods, distributionally robust optimization (DRO), from robust optimization literature, directly solves the OOD generalization problem by optimizing for the worst-case error over an uncertainty distribution set, so as to protect the model against the potential distributional shifts within the uncertainty set.
The uncertainty set is often constrained by moment or support conditions \cite{10.1287/opre.1090.0741,bertsimas2018data-driven}, $f$-divergence \cite{namkoong2016stochastic,duchi2018learning, sagawa2019distributionally} and Wasserstein distance \cite{esfahani2018data, SinhaCertifying, liu2021stable, DBLP:conf/wsc/BlanchetKMZ19, 9961896, NEURIPS2022_da535999,DBLP:journals/mor/BlanchetMZ22,DBLP:conf/nips/LiLBN22, DBLP:conf/aistats/LotidisBBL23}.
The objective function of DRO methods can be summarized as:
\begin{equation}
\label{equ:DRO-formulation}
	\arg\min_f \sup\limits_{Q\in \mathcal{P}(P_{tr})}\mathbb{E}_{X,Y\sim Q}[\ell(f(X),Y)]
\end{equation}
where $\mathcal{P}(P_{tr})$ is the distribution set lying around the training distribution $P_{tr}$ and $\ell(\cdot;\cdot):\mathcal{Y}\times\mathcal{Y}\rightarrow\mathbb{R}^+$ is the loss function.
Different DRO methods adopt different kinds of constraints to formulate the distribution set $\mathcal{P}(P_{tr})$ and correspondingly different optimization algorithms.
In this paper, we only introduce two typical DRO methods whose distribution sets are formulated by $f$-divergence and Wasserstein distance respectively, and for a more thorough introduction to DRO methods, one can refer to \cite{rahimian2019distributionally}.

\subsubsection{\texorpdfstring{$f$}{F}-Divergence Constraints}
\label{subsubsec:f-div}
The distribution set $\mathcal{P}(P_{tr})$ in $f$-divergence DRO \cite{duchi2018learning} is formulated as:
\begin{equation}
	\mathcal{P}(P_{tr}) = \{Q: D_f(Q\|P_{tr})\leq \rho\}
\end{equation}
where $\rho > 0$ controls the extent of the distributional shift, and $D_f(Q\|P_{tr}) = \int f(\frac{d Q}{d P_{tr}}) d P_{tr}$ is the $f$-divergence between $Q$ and $P_{tr}$.
Intuitively, if the potential testing distribution $P^{e_{test}}(X,Y) \in \mathcal{P}(P_{tr})$, DRO methods can achieve good generalization performance even if $P^{e_{test}}(X,Y) \neq P_{tr}(X,Y)$.
As for the optimization, a simplified dual formulation for the Cressie-Read family of $f$-divergence can be obtained.
\begin{lemma}[Optimization of $f$-divergence \cite{duchi2018learning}]
	For $f_k(t) = \frac{t^k-kt+k-1}{k(k-1)}$ and $k\in (1,+\infty)$, $k_*=k/(k-1)$, and any $\rho > 0$, we have for all $\theta \in \Theta$:
		\begin{equation}
			\mathcal{R}_k(\theta;P_{tr}) = \inf_{\eta\in\mathbb{R}}\left\{ c_k(\rho) \mathbb{E}_{P_{tr}}[(\ell(f(X),Y)-\eta)_+^{k_*}]^{\frac{1}{k_*}} + \eta \right\}
		\end{equation}
	where $c_k(\rho) = (k(k-1)\rho + 1)^{\frac{1}{k}}$.
\end{lemma}

\subsubsection{Wasserstein Distance Constraints}
Since the calculation of $f$-divergence requires the supports of two distributions to be the same while Wasserstein distance does not, the distribution set $\mathcal{P}(P_{tr})$ formulated by Wasserstein distance is more flexible.
Wasserstein distance is defined as:
\begin{definition}
	Let $\mathcal{Z} \subset \mathbb{R}^{m+1}$ and $\mathcal{Z} = \mathcal{X}\times\mathcal{Y}$ , given a transportation cost function $c: \mathcal{Z} \times \mathcal{Z} \rightarrow [0, \infty)$, which is nonnegative, lower semi-continuous and satisfies $c(z,z)=0$, for probability measures $P$ and $Q$ supported on $\mathcal{Z}$, the Wasserstein distance between $P$ and $Q$ is :
	\begin{equation}
		W_c(P, Q) = \inf\limits_{M \in \Pi(P,Q)} \mathbb{E}_{(z,z') \sim M}[c(z,z')]
	\end{equation}
	where $\Pi(P,Q)$ denotes the couplings with $M(A,\mathcal{Z})=P(A)$ and $M(\mathcal{Z},A)=Q(A)$ for measures $M$ on $\mathcal{Z}\times \mathcal{Z}$.
\end{definition}
Then the distribution set $\mathcal{P}(P_{tr})$ of Wasserstein DRO is formulated as:
\begin{equation}
	\mathcal{P}_c(P_{tr}) = \{Q: W_c(Q,P_{tr}) \leq \rho\}
\end{equation}
where the subscript $c$ denotes the transportation cost function $c(\cdot,\cdot)$.
However, Wasserstein DRO is difficult to optimize and works targeting different models and transportation cost functions have been proposed.
Wasserstein DRO for logistic regression was proposed by \citet{shafieezadehabadeh2015distributionally}.
\citet{SinhaCertifying} achieved moderate levels of robustness with little computational cost relative to empirical risk minimization with a Lagrangian penalty formulation of WDRO.
Recently, \citet{DBLP:conf/nips/LiLBN22} add martingale constraints to WDRO and derive tractable optimization for martingale DRO.
And \citet{9961896} incorporate geometric properties into DRO with geometric Wasserstein distance.

\subsubsection{Robustness Guarantees}
\label{subsubsec:robust}
Here we briefly review some theoretical results in DRO literature, including the relationship between regularization and robustness guarantees.

In order to demonstrate how the robust formulation (\ref{equ:DRO-formulation}) provides distributional robustness, several works establish the relationship between distributional robustness and regularization.
 For norm-based DRO methods, \citet{doi:10.1137/S0895479896298130} build the equivalence between the worst-case squared residual within a Frobenius norm-based distribution set and the Tikhonov regularization.
\citet{DBLP:journals/corr/abs-0811-1790} prove the equivalence between robust linear regression with feature perturbations and the Least Absolute Shrinkage and Selection Operator(LASSO).
\citet{DBLP:conf/icml/YangX13} and \citet{bertsimas2017characterization} make some further progress on this. 
For $f$-divergence-based DRO methods, \citet{duchi2021statistics} prove that the formulation (\ref{equ:DRO-formulation}) with distribution set $\mathcal{P}(P_{tr})=\mathcal{P}_{\rho,n}=\{p \in\mathbb{R}^n: p^T\mathbf{1}=1, p\geq 0, D_f(p\|\mathbf{1}/n)\leq \rho/n\}$ is a convex approximation to regularizing the empirical risk by variance.
For Wasserstein-based DRO methods, \citet{shafieezadehabadeh2015distributionally} investigate the Wasserstein DRO of logistic regression and show that the regularized logistic regression is one special case of it.
\citet{chen2018robust} also build the connection between the Wasserstein DRO of linear regression with $\ell_1$ loss function and regularization constraints on the regression coefficients.
\citet{shafieezadehabadeh2019regularization} and \citet{gao2017wasserstein} connect the Wasserstein DRO and regularizations in a unified framework.
\citet{DBLP:conf/nips/LiLBN22} prove that Wasserstein DRO is equivalent to Tikhonov regularization when exact martingale constraints are imposed.

As for the OOD generalization ability, in fact, the guarantees for OOD generalization of DRO methods naturally derive their formulation (\ref{equ:DRO-formulation}).
Since DRO methods directly optimize for the worst-case risk within the distribution set $\mathcal{P}(P_{tr})$, as long as the potential testing distribution $P_{te}\in\mathcal{P}(P_{tr})$, the OOD generalization ability is guaranteed.
Therefore, the remaining work is to provide the finite sample convergence guarantees, which ensure that the population-level objective $\sup_{Q\in\mathcal{P}(P_{tr})}\mathbb{E}_Q[\ell(f(X),Y)]$ can be optimized empirically with finite samples.
\citet{duchi2018learning} analyze the generalization bound of $f$-divergence-based DRO.
\citet{SinhaCertifying}, \citet{chen2018robust} and \citet{liu2021stable} also provide similar generalization bounds for Wasserstein DRO.
Also, \citet{levy2020large} come up with optimization methods for DRO of convex losses with conditional value at risk and $\mathcal{X}_2$-divergence uncertainty sets, which are suitable for large-scale applications.

\subsection{With Additional Information}
\label{sec:dro-env}
Although DRO methods could theoretically guarantee the out-of-distribution generalization ability when $P^{e_{test}}(X,Y)\in\mathcal{P}(P_{tr})$, there has been work questioning their real effects in practice.
 Intuitively, in order to achieve good OOD generalization ability, the potential testing distribution should be captured in the built distribution set.
 However, in real scenarios, to contain the possible true testing distribution, the uncertainty set is often overwhelmingly large, making the learned model make decisions with fairly low confidence, which is also referred to as the low confidence problem.
 Specifically, \citet{hu2016does} proved that in classification tasks, DRO ends up being optimal for the training distribution $P_{tr}$, which is due to the over-flexibility of the built distribution set. 
 And \citet{frogner2019incorporating} also pointed out the problem of overwhelmingly-large decision set for Wasserstein DRO.

 In order to overcome such a problem, \citet{DBLP:conf/wsc/BlanchetKMZ19} propose a data-driven way to select the transportation cost function. 
\citet{frogner2019incorporating} propose to further restrict the distribution set with a large number of unlabeled data. 
\citet{9961896} notice that in real scenarios, different covariates may be perturbed in a non-uniform way, and form a more reasonable distribution set according to the stability of covariates across environments.
\citet{duchi2019distributionally} assume that $P(Y|X)$ stays invariant and propose to only perturb the marginal distribution $P(X)$ to deal with covariate shifts.
Though some meaningful attempts, how to incorporate additional information to form a more reasonable distribution set is also an open problem.
We refer readers to \citep{DBLP:journals/corr/abs-1908-05659} for a more comprehensive survey.

\section{Theoretical Connections}
\label{sec:theory}
For branches of methods for OOD generalization, there are some inherent connections among them. 
In this section, we will demonstrate the connections among causal learning methods, distributionally robust optimization (DRO) methods, and stable learning methods, which may benefit the understanding of OOD generalization methods.

\subsection{DRO and Causality}
\label{sec:dro-causal}
Recall that DRO methods aim to optimize the worst-case error over a pre-defined distribution set, so as to protect the learned model from potential distributional shifts, which often take the form of:
\begin{equation}
	\arg\min_f \sup\limits_{Q\in\mathcal{P}(P_{tr})}\mathbb{E}_{X,Y\sim Q}[\ell(f(X),Y)]
\end{equation}
where $\mathcal{P}(P_{tr})$ is the distribution set built around the training distribution $P_{tr}$.
Although in DRO literature, $\mathcal{P}(P_{tr})$ is often characterized by $f$-divergence or Wasserstein distance, different choices of $\mathcal{P}(P_{tr})$ will render DRO equivalent to causal inference in the structural equation model (SEM) context\cite{8439889}, which shows the inherent relationship between causality-based methods and DRO methods.
Taking linear equation models for example, suppose we have a directed acyclic graph $G=(V,E)$ with $p$ nodes $V=\{1, \dots, p\}$ and correspondingly a $p$-dimension random variable $Z$, then the training distribution is determined by the structural causal model (SCM) as:
\begin{equation}
	Z = BZ + \epsilon
\end{equation}
where $Z=(X,Y)\in\mathbb{R}^p$ is the random variable of interest, $B\in\mathbb{R}^{p\times p}$ is the coefficient matrix and $\epsilon \sim P_{\epsilon}$ the random noise.
We will show that finding causal coefficients for predicting $Y$ can be reformulated as performing DRO on interventional distribution set, including do-interventional and shift-interventional distributions.

Do-interventions on variables $S \subseteq V$ can be formulated as:
\begin{align}
	Z_k = (BZ)_k + \epsilon\quad \text{for $k\not\in S$};\quad 
	Z_K = A_k\quad \text{for $k\in S$}
\end{align}
where $A\in\mathbb{R}^p$ and the value of the do-intervention on variable $k\in S$ is $A_k$.
Then the error distribution $P_\epsilon$, coefficient matrix $B$, intervention set $S \subseteq V$ and intervention value $A\in\mathbb{R}^p$ induces a distribution for a random variable $Z(A, S)$, denoted as $Z(A,S) \sim P_{A,S}^{(\text do)}$.
And the corresponding do-interventional distribution set can be formulated as $\mathcal{P}^{(\text do)} = \{P_{A, V/\{p\}}^{(\text do)}: A\in\mathbb{R}^p\}$.
Analogously to do-interventions, the shift-interventions is defined as:
\begin{equation}
	Z = BZ + \epsilon + A
\end{equation}
where $A\in\mathbb{R}^p$ is the shift direction, and the induced distribution is denoted as $Z(A)\sim P_A^{(\text{shift})}$ and the shift-interventional distribution set can be formulated as $\mathcal{P}^{(\text{shift})} = \{P_A^{(\text{shift})}: A_p=0\}$.

When performing DRO on $\mathcal{P}^{(\text do)}$ or $\mathcal{P}^{(\text{shift})}$, causal coefficients can be obtained~\cite{8439889} since
\begin{equation}
	\min_\theta\sup\limits_{Q\in \mathcal{P}^{(\text{do})}}\mathbb{E}[\ell(f_\theta(X),Y)] = \left\{
             \begin{array}{lr}
             \infty, &\text{if $\theta\neq \theta_{\text{causal}}$}  \\
             \text{Var}(\epsilon_p), &\text{if $\theta= \theta_{\text{causal}}$}  
             \end{array}
\right.
\end{equation}
and 
\begin{equation}
	\min_\theta\sup\limits_{Q\in \mathcal{P}^{(\text{shift})}}\mathbb{E}[\ell(f_\theta(X),Y)] = \left\{
             \begin{array}{lr}
             \infty, &\text{if $\theta\neq \theta_{\text{causal}}$}  \\
             \text{Var}(\epsilon_p), &\text{if $\theta= \theta_{\text{causal}}$}  
             \end{array}
\right.
\end{equation}
which reveals that causal inference can also be viewed as a special case of distributional robustness.

\subsection{Stable Learning and Causality}
\citet{xu2021stable} theoretically analyze stable learning algorithms through the lens of feature selection and connect them with causality.
They first prove that these algorithms could identify a certain set of variables defined as follows.

\begin{definition} [Minimal stable variable set \cite{xu2021stable}] \label{defn:stable-set}
    A minimal stable variable set of $Y$ under distribution $P$ is any subset $\mathbf{S}$ of $\mathbf{X}$ for which
    \begin{equation} \label{eq:sstable-set}
        \mathbb{E}_P[Y | \mathbf{S}] = \mathbb{E}_P[Y | \mathbf{X}],
    \end{equation}
    and none of $\mathbf{S}$'s proper subsets satisfies \Cref{eq:sstable-set}.
\end{definition}

They theoretically show that the minimal stable variable set is minimal and optimal to deal with covariate-shift generalization for common loss functions \cite[Theorem 3]{xu2021stable}. As a result, the effectiveness of stable learning algorithms on covariate-shift generalization could be proved.

Furthermore, they show that the minimal stable variable set is a subset of the Markov boundary \cite{pearl2014probabilistic}. Markov boundary discovery is generally challenging because traditional methods \cite{aliferis2010locala,aliferis2010localb} are always based on the conditional independence test, which is a particularly difficult hypothesis to test for \cite{shah2020hardness}. As a result, stable learning algorithms could help discover the Markov boundary to some extent, which can be of independent interest.

\section {Evaluation for OOD Generalization}
\label{sec:benchmark}
To promote the research of OOD generalization, it is of vital importance to evaluate the OOD generalization performances of different algorithms.
In this section, we summarize the datasets commonly used in literature.

\begin{table*}[t]
    \centering
    \caption{Commonly used image datasets for OOD generalization. Shift type denotes the type of distributional shifts, and the mixed type in image type means that there are both real and unreal images.}
    \resizebox{\textwidth}{24mm}{
    \begin{tabular}{|l|c|c|c|c|c|c|c|}
    \hline
        \multirow{2}*{Image Data} & \bf ImageNet-Variant &\bf Colored MNIST &\bf MNIST-R &\bf Waterbirds &\bf Camelyon17 &\bf VLCS &\bf PACS\\
        & \cite{hendrycks2021natural,hendrycks2019benchmarking,hendrycks2020many} & \cite{arjovsky2019invariant} & \cite{ghifary2015domain} & \cite{sagawa2019distributionally} & 
        \cite{bandi2018detection} & \cite{fang2013unbiased} & \cite{li2017deeper}\\
    \hline
        \# Domains & - & 3 & 6 & 2 & 5 & 4 & 4 \\
    \hline
        \# Categories & - & 2 & 10 & 2 & 2 & 5 & 7 \\
    \hline
        \# Examples & - & - & 6k & 4.8k & 450k & 2.8k & 9.99k \\
    \hline
        Shift Type & Adversarial Policy & Color & Angle & Background & Hospital & Data Source & Style \\
    \hline
        Image Type& Mixed Type & Digits & Digits & Birds & Tissue Slides & Real Objects & Mixed Type\\
        \hline
        \hline
        \multirow{2}*{Image Data} & \bf Office-Home &\bf DomainNet & \bf iWildCam &\bf FMoW &\bf PovertyMap & 
        \bf NICO & \bf NICO++ \\
        & \cite{venkateswara2017deep} & \cite{DBLP:conf/iccv/PengBXHSW19} & \cite{beery2021iwildcam} & \cite{christie2018functional} & 
        \cite{yeh2020using} &
        \cite{he2021towards} & 
        \cite{zhang2022nico++}\\
    \hline
        \# Domains & 4 & 6 & 323 & 16 $\times$ 5 & 23 $\times$ 2 & 
        188 & 810 \\
    \hline
        \# Categories & 65 & 345 & 182 & 62 & Real Value &
        19 & 80 \\
    \hline
        \# Examples & 15.5k & 570k & 200k & 500k & 20k &
        25k & 230k \\
    \hline
        \multirow{2}*{Shift Type} & \multirow{2}*{Style} & \multirow{2}*{Style} & \multirow{2}*{Location} & Time,  & Country,  & \multicolumn{2}{c|}{Background, Attribute, Action,} \\
        & & & & Location &  Urban/Rural & \multicolumn{2}{c|}{View and Co-occurring Object} \\
    \hline
        Image Type & Mixed Type & Mixed Type & Real Animals & Satellite & Satellite & \multicolumn{2}{c|}{Real Objects} \\
    \hline
    \end{tabular}}
    \label{tab:bk}
\end{table*}

Datasets can be classified according to different criteria (e.g., synthetic data and real-world data; tabular data, image data, and language data).
And researchers from different fields utilize different kinds of datasets, for example, statistical machine learning often uses synthetic \& tabular data, while computer vision researchers often use real-world \& image data.
As for the OOD generalization, it is necessary to involve distribution shifts to evaluate the generalization ability of different approaches.
In line with recent works, we present a comprehensive overview of datasets and evaluation metrics of OOD generalization.

\subsection{Synthetic Data}
Synthetic data are important for simulating explainable and controllable distribution shifts.
\citet{aubin2021linear} find that recent OOD generalization methods perform poorly on some simple low-dimensional linear problems.
This demonstrates the need for such simple but challenging data, which could precisely reflect to what extent an algorithm could resist certain kinds of distribution shifts. 

In this section, we introduce three typical synthetic data generation mechanisms, with which one can simulate certain kinds of distribution shifts to various degrees and evaluate the generalization ability of different algorithms.
Throughout these mechanisms, covariates $X$ are divided into two groups as $X=[S,V]^T$, corresponding to the stable and unstable/spurious parts, \emph{i.e.} $P(Y|S)$ remains invariant across distributions and $P(Y|V)$ is perturbed to bring distributional shifts.

\paragraph{Unobserved Confounders}\quad 
Confounding bias is one of the most sources of distribution shifts~\citep{pearl2009causality, subbaswamy2018counterfactual, arjovsky2019invariant}, where the unstable covariates $V$ are related to target $Y$ owing to the unobserved confounder $C$.
Here we present the data generation process proposed by \citet{subbaswamy2018counterfactual}:
\begin{align}
	V = W^e_VC + \epsilon_V,\quad Y = W_S^TS + W_cC + \epsilon_Y, 
\end{align}	
where $C$ is the unobserved confounder. 
Coefficient $W_V^e$ controls the relationship between $V$ and $Y$, and one can change $W_V^e$ across environments to simulate distribution shifts.

\paragraph{Selection Bias}\quad 
\citet{kuang2020stable} propose a selection bias mechanism, and similar settings are also adopted in \cite{liu2021stable, hrm, DBLP:journals/corr/abs-2212-00992}.
In this setting, $P(Y|V)$ is perturbed with selection bias.
The data generation process is as follows:
	\begin{align}
		Y = f(S)+\epsilon = \theta_S^TS + \beta S_1\cdot S_2\cdot S_3 + \epsilon,
	\end{align}	
and the sample selection probability $\hat{P}(X)$ of each data point follows:
	\begin{align}
		\hat{P}(X) = \prod_{v_i \in V}|r|^{-5*|f(S) - \text{sign}(r)*v_i|}.
	\end{align}	
$|r|>1$ is the bias factor to control the strength of distribution shifts.
The larger value of $|r|$ brings the stronger spurious correlation between $V$ and $Y$, and $r \ge 0$ means positive correlation and vice versa.

\paragraph{Regression from Causes and Effects}
\citet{arjovsky2019invariant} and \citet{hrm} introduce an anti-causal mechanism to change $P(Y|V)$.
In this setting, the data generation process is defined as:
\begin{align}
	Y = W_SS + \epsilon_Y,\quad V = W_V^eY + \epsilon_V^e
\end{align}	
where the coefficient $W_V^e$ and $\epsilon_y, \epsilon_V^e$ control the relationship between $V$ and $Y$.
Intuitively, larger $\epsilon_Y$ and smaller $\epsilon_V^e$ will make the model easier to utilize $V$ for prediction, making OOD generalization more challenging. 

There are various synthetic data generation mechanisms in literature, and one can refer to \citep{aubin2021linear, pmlr-v119-sagawa20a, NEURIPS2022_da535999} for more synthetic settings.

\subsection{Real-World Data}
Although synthetic data could reflect the generalization ability of different approaches, they are difficult to generate complicated data (e.g., image/language data), and whether the simulated shift patterns correspond with real-world scenarios remain unclear.
To demonstrate the practical value of OOD generalization methods, it is necessary to involve real-world datasets for evaluation. 
Here, we describe several typical real-world (and pseudo-real) benchmarks used in OOD generalization literature, including image, tabular, language, graph, and code data.

\paragraph{Image Data} 
With the rapid development of computer vision, a number of image datasets have been released.
According to the flexibility of customizing distribution shifts, we classify them into three categories, namely pseudo-real shifts, static natural shifts, and controllable natural shifts.
A summary of these datasets is shown in \Cref{tab:bk}.\\
\textsc{(a) Pseudo-Real Shifts}.\quad
For image datasets not designed for OOD generalization, some synthetic transformations are added to introduce distribution shifts.
The most typical ones, including ImageNet~\cite{deng2009imagenet} variants (e.g. ImageNet-A~\cite{hendrycks2021natural}, ImageNet-C~\cite{hendrycks2019benchmarking}, ImageNet-R~\cite{hendrycks2020many}) adopt data selection mechanisms or perturbations to generate testing data with distribution shifts.
Others, typified by MNIST~\cite{lecun1998gradient} variants (e.g. Colored MNIST~\cite{arjovsky2019invariant}, 
    MNIST-R~\cite{ghifary2015domain}), simulate different environments by coloring or rotating original images.
And Waterbirds~\cite{sagawa2019distributionally} introduces spurious correlations between bird categories and backgrounds.
These datasets make it available for preliminary study and evaluation of OOD generalization approaches.\\
\textsc{(b) Static Natural Shifts}.\quad 
Recently, there are a few datasets supporting OOD generalization validation, which mainly involve natural shifts, e.g., spatial and temporal shifts.
Widely used in domain generalization, PACS~\cite{li2017deeper} and Office-Home~\cite{venkateswara2017deep} design environments according to image styles, and VLCS~\cite{fang2013unbiased} and  iWildCam~\cite{beery2021iwildcam} directly uses data sources as environments.
Besides, Camelyon17~\cite{bandi2018detection} contains tissue slides sampled and post-processed in different hospitals and 
DomainNet~\cite{zhao2019multi} extends PACS to a larger scale, consisting of more domains and categories.
Recently, \citet{koh2021wilds} collect several datasets together and produce Wilds as a benchmark for OOD generalization.
And \citet{DBLP:conf/nips/YaoCC0KF22} curate Wild-Time to reflect temporal distribution shifts in various real-world applications.\\
\textsc{(c) Controllable Natural Shifts}. \quad 
Recently, there are datasets enabling more flexible and controllable ways to simulate distributional shifts, typified by NICO~\cite{he2021towards} and NICO++~\cite{zhang2022nico++}.
NICO elaborately selects visual contexts with various types, including background, attribute, view and etc.
With diverse contexts, NICO could simulate different types of natural shifts, and with a balanced sample size in each context, different degrees of distribution shifts could be easily produced.
As an extended version of NICO, NICO++ splits domains into common domains (shared by all categories) and unique domains (for each category). 
For each category, NICO++ contains 10 common domains and 10 unique domains, supporting both typical DA and OOD generalization settings with flexible and controllable shifts. 
Besides, FMoW\cite{christie2018functional} collects satellite images of buildings or land with tokens at different times and regions, and PovertyMap\cite{yeh2020using} contains images of an urban or rural area from disjoint sets of countries.

\paragraph{Tabular Data}
Tabular data widely exist in real-world high-stake applications, including economics, health care, and so on.
Therefore, it is important to deal with natural distribution shifts in tabular data.
House sales price dataset\footnote{https://www.kaggle.com/c/house-prices-advanced-regression-techniques/data} considers temporal shifts in price prediction and is used in \cite{shen2020stable, liu2021stable, hrm}.
And demographic shifts are considered in Adult\footnote{https://archive.ics.uci.edu/dataset/2/adult}, BRFSS\footnote{https://www.cdc.gov/brfss/}, COMPAS\footnote{https://www.kaggle.com/datasets/danofer/compass} datasets.
Spatial shifts are considered in ACS datasets~\cite{DBLP:conf/nips/DingHMS21}, which contains data from 51 US states.
Recently, \citet{DBLP:journals/corr/abs-2307-05284} propose WHTSHIFT, an empirical testbed with curated real-world shifts, where the type of shift is specified for each of the 22 settings.

\paragraph{Others} 
OGB-MolPCBA~\cite{ren2020open} collects molecular graphs in over 100,000 scaffolds and formulates a molecular property prediction task across different scaffolds.
CivilComments~\cite{borkan2019nuanced} and Amazon\cite{ni2019justifying} gather the individual comments of different users and distinctive groups (e.g. male and female).
GLUE-X~\citep{yang2023gluex} provides a unified benchmark for evaluating OOD robustness in NLP models.
Towards auto-engineering, Py150~\cite{raychev2016probabilistic} contains codes from 8,421 git repositories for code completion generalization.

\subsection{Empirical Findings}
Recently, there are works investigating the OOD generalization performances in a purely empirical way, which provide valuable insights.
\citet{DBLP:conf/iclr/GulrajaniL21} find that the real effects of domain generalization approaches are relatively weak on real-world image data.
\citet{pmlr-v139-miller21b} empirically show that the OOD performance is strongly correlated with in-distribution performance on image data for a wide range of deep models and distribution shifts.
\citet{yang2023change} release a comprehensive benchmark of 20 algorithms with 12 real-world datasets in vision, language, and healthcare domains, and they empirically study the relationships between different evaluation criteria.
And \citet{DBLP:journals/corr/abs-2307-05284} empirically validate the prevalence of $Y|X$ shifts in real-world tabular data, where the accuracy-on-the-line phenomena do not hold.
This addresses the importance of specifying the shift patterns on tabular data, and they release a benchmark with 22 specified distribution shift patterns.

\section{Implications for fairness and explainability}
\label{sec:fairness}

\subsection{Fairness}
Nowadays, fairness issues have raised great concerns in decision-making systems such as loan applications \cite{mukerjee2002multi}, hiring processes \cite{rivera2012hiring}, criminal justice \cite{larson2016we}, personalized pricing \cite{xu2022regulatory}, and online markets \cite{xu2022product}. 
Poorly designed algorithms tend to amplify data bias, resulting in discrimination against specific subgroups of individuals based on their inherent characteristics, which are often named sensitive attributes in fairness problems. 
Many works define their fairness and propose corresponding fair algorithms, from which the definition of fairness can be divided into three types: individual fairness \cite{dwork2012fairness,yurochkin2021sensei}, group fairness \cite{hardt2016equality, kearns2018preventing, xu2020algorithmic}, and causality-based fairness notions \cite{kilbertus2017avoiding, chiappa2019path}. 
However, different fairness notions are in conflict \cite{kleinberg2016inherent}. Methods that mitigate unfairness in the algorithms fall under three categories: pre-processing \cite{wang2019repairing, feldman2015certifying, kamiran2012data}, in-processing \cite{zafar2017fairnessa, zafar2017fairnessb, agarwal2018reductions}, and post-processing \cite{hardt2016equality} algorithms. 

Fairness has recently been linked to OOD issues, according to \citet{creager2020environment}. 
Generally speaking, subgroups split by sensitive attributes in fairness literature correspond to environments in OOD literature. 
Following that, both areas need to specify learning objectives with respect to the subgroups/environments. In fairness literature, the learning objectives represent context-specific fairness notions, while in OOD literature, the learning objectives should be designed according to invariance assumptions. Similar learning objectives could be adopted in both areas. 
For example, objectives similar to fairness criterion equalized odds \cite{hardt2016equality} are adopted in OOD literature \cite{li2018deep, ahmed2021systematic} to deal with simplicity bias \cite{shah2020pitfalls}. 
The learning objective of IRM \cite{arjovsky2019invariant} is also similar to calibration in fairness literature \cite{chouldechova2017fair}. 
Meanwhile, classical approaches from OOD literature could be applied to address fairness issues. 
Fair representation learning methods \cite{edwards2016censoring,xu2020algorithmic,zhao2020conditional} originated from domain adaptation (DA) methods \cite{ben2010theory, ganin2016domain}. 
When sensitive attributes are unknown, DRO and adversarially learning were introduced in fairness literature \cite{hashimoto2018fairness, lahoti2020fairness, duchi2020distributionally} to obtain a distributionally robust predictor and ensure the worst subgroup performance. \cite{kearns2018preventing, hebert2018multicalibration, kim2019multiaccuracy} also adopt adversarially learning methods to ensure all computationally identifiable subgroups are treated equally. 
As a result, Pursuing OOD could be considered as pursuing fairness concerning the subgroups/environments if the invariance assumption adopted for OOD could be viewed as a fairness notion.

In addition to considering subgroups as environments, \cite{mandal2020ensuring}  investigate another scenario in which the environment is a separable variable. They studied fair classifiers that are robust to perturbations in the training distribution and devised a DRO-like method to reach their goal. 
Fair and robust learning is also applied in \cite{roh2020fr}. These works differ from the works listed in the last paragraph in that fairness and robustness are two objectives here whereas the aforementioned works consider them the same.

\subsection{Explainability}
Explanation methods can be generally divided into post hoc analyses and model-based methods~\cite{murdoch2019definitions}. There exist several works in both directions. Post hoc analyses usually explain a black-box model by calculating feature importance \cite{adebayo2018sanity}. Typical methods include gradient-based \cite{selvaraju2017grad,ramanishka2017top,dabkowski2017real}, influence function \cite{koh2017understanding}, and Shapley values \cite{lundberg2017unified}. Model-based explanation methods often adopt simpler hypotheses such as linear regression \cite{friedman2001elements}, LASSO \cite{tibshirani1996regression}, generalized additive models \cite{hastie1990generalized}, decision trees \cite{friedman2001elements}, and rule-based methods \cite{friedman2008predictive,letham2015interpretable}.

Causality \cite{pearl2009causality} has recently been introduced to model explanation, especially in deep learning methods. Traditional deep-learning algorithms are rarely used in high-stakes applications due to their lack of explainability. Causality could provide a way to shed light on the explainability of deep learning. For example, several works adopt causality to explain deep models in textual and visual explanation \cite{alvarez2017causal,anne2018grounding, goyal2019counterfactual}. Furthermore, the Causal And-Or Graph was proposed in robotics \cite{xiong2016robot} and object tracking \cite{xu2018causal} to build explainable algorithms with the knowledge of causality. \citet{kim2017interpretable} also applied a causal filtering step in self-driving automobile problems.
\begin{equation} \label{eq:oodcausalexplain}
	\text{OOD} \Leftarrow \text{Causality} \Rightarrow \text{Explainability}.
\end{equation}

Actually, causality is the crux for both OOD generalization and explainability as shown in \Cref{eq:oodcausalexplain}. The models will have good OOD generalization performance and explainability simultaneously if they utilize the causal relationship between the features and the outcomes. Hence, explainability would be a side product when pursuing OOD generalization with causality.


\section{Conclusion and Future Directions}
\label{sec:conclusion}
Out-of-Distribution (OOD) generalization problem has aroused much research attention recently and is critical for the deployment of machine learning algorithms.
In this paper, we systematically and comprehensively review the definition, the main branches of methods, theoretical connections among different methods, and the datasets of the OOD generalization problem. 
Finally, we list several potential challenges in OOD generalization and we hope they could inspire future research on OOD generalization problem.

\paragraph{Theoretical characterization}
Although growing popular recently, the theoretical characterization of a learnable OOD generalization problem remains vague in recent literature.
Characterizing the learnability of a problem is a basic question in machine learning.
Though previous research efforts have been made in $i.i.d.$ setting, the learnability is difficult to define and analyze under distributional shifts, since it is impossible to enable models to generalize to arbitrary and unknown distributions.
Therefore, in OOD generalization problem, figuring out what kind of distributional shifts should be taken into consideration is critical for the analysis of learnability.
There is very little exploration~\cite{Ye2021a} on this and more research efforts need be paid on this.

\paragraph{Demands for environments}
Multiple training environments are required for the majority of OOD generalization methods, while in practice modern datasets are often assembled by merging data from multiple sources without keeping source labels. This greatly restricts the deployment of OOD generalization methods in real scenarios.
Therefore, it is more practical and realistic that we only have access to one training environment with latent heterogeneity.
Recently, while there are some works~\cite{creager2020environment, hrm} try to leverage the latent heterogeneity and relax the demands for environments, how to explore and utilize the latent heterogeneity inside data is critical for the deployment of OOD generalization methods and is a promising future direction.

\paragraph{Reasonable evaluations}
Although the evaluation criteria for classic machine learning algorithms under $i.i.d.$ assumption are well-developed, including testing data, model selection mechanisms, and so on, they cannot directly be deployed to OOD scenarios. 
Since the testing distribution is both different and unknown from the training, how to design fair and realistic experimental settings remains a challenging problem.
Further, the model selection mechanism also matters, since the choice of validation data is non-trivial in OOD scenarios, and \citet{DBLP:conf/iclr/GulrajaniL21} also demonstrate that domain generalization algorithms without a model selection strategy are incomplete.
Also, \citet{DBLP:conf/iclr/GulrajaniL21} notice that the real effects of many domain generalization methods are weak, which indicates that existing evaluation criteria are inadequate to validate an OOD generalization algorithm.
And \citet{rethinking} reflect on the evaluation protocol of domain generalization. They investigate and demonstrate the test data information leakage from pre-trained weights and a single test environment in the current evaluation protocol. 
Therefore, it is critical for the community to develop more reasonable evaluation criteria for OOD generalization.

\paragraph{Incorporation of Pre-Trained \& Large Language Models}
Recently, there has been a surge in the development of large language models (or pre-trained models), such as BERT~\cite{devlin2018bert}, GPT-3~\cite{DBLP:conf/nips/BrownMRSKDNSSAA20}, SimCLR~\cite{DBLP:conf/icml/ChenK0H20}, StableDiffusion~\citep{rombach2022high}, ChatGPT\footnote{https://openai.com/blog/chatgpt}, GPT-4\footnote{https://openai.com/gpt-4}. 
These models propose an approach of initially pre-training on large-scale datasets, followed by fine-tuning or directly deploying on downstream tasks.
Since it's inevitable to encounter distribution shifts between downstream tasks and pre-training datasets, devising efficient pre-trained methods with strong OOD generalization ability becomes critical. 
Alternatively, the integration of pre-trained methods to enhance OOD generalization performance is also a promising direction for future exploration.
Furthermore, it is becoming more important to evaluate the OOD generalization ability of large language models in deployment~\citep{yang2023outofdistribution, wang2023robustness}.

\newpage
\bibliography{tkde}
\bibliographystyle{abbrvnat}

\end{document}